\documentclass[final,5p,times,twocolumn]{elsarticle}

\usepackage{amssymb}
\usepackage{amsmath}
\usepackage{natbib}
\usepackage{url}
\usepackage{booktabs}
\usepackage{tabularx}

\journal{Future Generation Computer Systems}

\begin{document}

\begin{frontmatter}

\title{Multimodal Deep Learning for Flood Detection: Integrating Water Level Sensors, Rainfall, Cameras, and Satellite Imagery\tnoteref{t1}}
\tnotetext[t1]{Accepted manuscript. Published in \emph{Future Generation Computer Systems} 188 (2027) 108870, \url{https://doi.org/10.1016/j.future.2026.108870}. \copyright~2026. This manuscript version is made available under the CC-BY-NC-ND 4.0 license \url{https://creativecommons.org/licenses/by-nc-nd/4.0/}.}

\author[ttu]{Ipshita Ahmed Moon}
\ead{imoon42@tntech.edu}

\author[ttu]{Jack Barbieri}
\ead{jmbarbieri42@tntech.edu}

\author[ttu]{Alfred Kalyanapu}
\ead{akalyanapu@tntech.edu}

\author[ttu]{Susmit Shannigrahi\corref{cor1}}
\ead{sshannigrahi@tntech.edu}

\cortext[cor1]{Corresponding author}

\affiliation[ttu]{
    organization={Tennessee Technological University},
    city={Cookeville},
    state={Tennessee},
    country={USA}
}

\begin{abstract}
Flood detection systems typically treat hydrometers, rainfall sensors, cameras, and satellite imagery as separate streams, which delays alerts and destabilizes stage transitions whenever a single sensor is noisy or unavailable. This work presents a multimodal forecasting framework fusing all four modalities into one temporal model, converting camera and satellite imagery into continuous depth signals learned alongside physical sensor readings. Unlike formulations that classify the current flood stage from concurrent readings, the system forecasts the stage several steps ahead from only past observations, removing the circularity between what the model observes and what it predicts. Six architectures (LSTM, GRU, TCN, Transformer, Random Forest, XGBoost) are compared under identical conditions across three forecast horizons, with early- and late-fusion variants and five-seed statistical testing throughout. Multimodal fusion significantly outperforms the best single-sensor baseline at short and medium horizons, though this advantage does not hold at every horizon, a finding reported directly rather than hidden behind a best-case result. A lightweight, rule-based reasoning layer requiring no training data reduces stage oscillations by 25\% while improving macro F1, evaluated directly against real field data from a flood-prone site, not simulation alone. That evaluation surfaces a specific limitation: tree-based architectures, despite strong aggregate accuracy, fail to detect severe flood events at longer horizons, a blind spot invisible in the aggregate score but confirmed on genuine field measurements. A robustness evaluation shows the system tolerates missing readings far better than corrupted ones. These results demonstrate a forecasting system tested under realistic, imperfect conditions, not idealized simulation alone.
\end{abstract}

\end{frontmatter}

\section{Introduction} \label{sec:intro}

When a small creek rises fast enough to flood a road crossing within an hour, the sensors watching it rarely agree on what is happening. A rain gauge upstream may already show a burst of precipitation while the water level at the crossing itself has not moved yet. A camera trained on the crossing might show water rising in real time while the nearest hydrometer, damaged or simply absent, reports nothing at all. Treated as independent alarms, these signals produce exactly the kind of system this paper argues against: one that is silent when a sensor fails, unstable when two sensors briefly disagree, and slow to act because no single source can confirm a flood on its own. Flooding is the most frequent and economically destructive natural hazard worldwide, and the fragmentation of how it is monitored, hydrometers, rainfall gauges, cameras, and satellite imagery each analyzed on their own, is a large part of why detection systems still produce delayed alerts and unreliable stage transitions.

The difficulty is not that any one modality is uninformative, but that each is informative about a different part of the same event. Rainfall sensors register the earliest signal but are prone to false alarms, since not all rainfall produces flooding. Water-level sensors are the most direct evidence of an active flood but by definition confirm it only after it has already started. Satellite imagery adds spatial context no point sensor can offer, at the cost of temporal resolution too coarse for real-time alerting. None of this is solved by simply collecting all four signals; it requires a model that learns how they relate to each other over time, and a way to keep that model's output stable when the underlying signals are themselves noisy or incomplete.

This paper presents a multimodal forecasting framework built around that idea. Hydrometer depth, rainfall, camera-derived depth, and satellite-derived depth are fused into a single temporal sequence model, with camera and satellite imagery first converted into continuous depth time series so they can be learned jointly with the physical sensor streams rather than processed by separate pipelines. The system is trained to forecast the flood stage several steps ahead of its most recent observation, rather than classify the stage at the same timestep as its inputs, which removes a circularity between inputs and labels that affects the same-timestep formulation used in prior work. A rule-based reasoning layer sits downstream of the forecasting model and suppresses short-lived, noise-driven stage changes without constraining how large a single genuine transition can be, so that a flash flood's abrupt onset is not mistaken for an artifact and smoothed away.

Two questions follow directly from this design, and both are answered empirically rather than assumed. First, does fusing four modalities actually outperform relying on the best single sensor, and does that advantage hold across every forecast horizon or only some? Six architectures, LSTM, GRU, TCN, Transformer, Random Forest, and XGBoost, are compared under identical conditions, with both early- and late-fusion variants and statistical significance testing across five random seeds, specifically to answer this without relying on a single best-case configuration. Second, does any of this transfer beyond the simulated data it was developed on? The system is evaluated directly against real field measurements, real rainfall, real sensor depth, and real camera imagery, collected at a flood-prone site, without any retraining, and the results are reported alongside the simulated ones rather than in place of them, including a limitation that only shows up on real data: a systematic failure of tree-based architectures to detect severe flood events at longer horizons, invisible in aggregate accuracy but confirmed in the field.

The remainder of the paper develops this argument in order. Section~\ref{sec:related} situates the approach against prior single- and dual-modality flood detection work. Section~3 defines the forecasting task and the dataset, architecture, and reasoning-layer design. Section~4 specifies the experimental protocol, and Section~5 reports results for the fusion comparison, the reasoning layer, robustness under degraded sensor conditions, and the real-world evaluation in turn. Section~6 discusses what these results imply for deployment and where the current design still falls short, and Section~7 concludes.

\section{Background and Related Work}
\label{sec:related}

\subsection{Flood Risk and Monitoring Challenges}

Flooding is the most frequent and economically destructive natural hazard worldwide,
causing cascading damage to infrastructure, agriculture, and human life
\cite{carter2018,paprotny2018,tabari2020,easterling2017,kunkel2012}.
The NOAA National Centers for Environmental Information (NCEI) estimates that
large-scale flood events in the United States over the past four decades resulted
in losses exceeding \$84.6 million USD per week and 738 casualties
\cite{smith2024}.
Critically, while headline-grabbing large floods attract public and governmental
attention, smaller-scale flooding in semi-urban and rural areas is often
under-reported yet equally devastating \cite{bronstert1995,spitalar2014}.
Climate change is both intensifying and spatially expanding flood risk,
exposing previously unaffected regions to extreme inundation events
\cite{tabari2020,dullo2021}.

Despite the growing urgency, continuous flood monitoring infrastructure
remains expensive and sparse.  A single standard USGS stream-gauge costs
between \$25{,}000 and \$40{,}000 to install and maintain \cite{normand2020},
putting dense coverage out of reach for most small and rural communities.
Commercial alternatives, such as satellite imaging, weather radar, IoT sensor
networks, and hydrological models, each provide partial views of the problem
but share a common structural limitation: they operate as \emph{siloed},
single-modality systems.  Satellite imagery suffers from cloud-cover
interference and low revisit frequency \cite{portalescjulia2023,notti2018};
radar measures precipitation indirectly \cite{berne2004}; IoT sensor deployments
require extensive infrastructure and maintenance \cite{9239693};
and crowd-sourced social-media signals are noisy and unverified.  Table~\ref{tab:comparison} summarises these limitations
and motivates the integrated, multimodal approach we pursue in this work.

\subsection{AquaCam: A Low-Cost Camera-Based Water-Level Sensor}

Our prior work introduced AquaCam, a deploy-anywhere, low-cost water-level
detection instrument built around a Raspberry Pi and an infrared camera
\cite{latham2025aquacam}.
The sensor captures periodic images of a stream, uses a YOLOv8
convolutional neural network (CNN) trained on a virtual staff-gauge bounding
box to distinguish water from background, and derives water height from the
visible fraction of the gauge within that bounding box.  Training on
approximately 2{,}000 labelled images over 100 epochs yielded a precision
of 96\% and recall of 95.3\% (F1 $\approx$ 0.96) on the initial test set.

A key contribution of that work was demonstrating \emph{transfer inference}:
a model trained exclusively at a Little Creek deployment in Tennessee was
applied without retraining to a geographically distinct site and achieved
accuracy~87\%, recall~82\%, and F1~90\%.  This result confirmed the
practical generalisability of the approach and established the foundation
for deploying a network of such sensors.  The sensor is solar-powered,
transmits data via 4G~LTE/Wi-Fi, and has accumulated more than 15{,}000
stream images spanning diverse lighting, weather, and seasonal conditions
\cite{kalyanapu2023lowcost}.

AquaCam provides an important but deliberately narrow view of flood risk:
it measures a single water-level time series at a fixed location.  The
results in~\cite{latham2025aquacam} explicitly identified the absence of
complementary data streams (rainfall, regional weather forecasts, and
satellite inundation signals) as the primary limitation for moving from
\emph{detection} to \emph{forecasting}.  This observation directly motivates
the multimodal framework developed in the present paper.

\subsection{Single-Modality Machine-Learning Approaches and Their Limitations}

A substantial body of work has demonstrated that machine learning can extract
useful flood-warning signals from individual sensor streams, yet each unimodal
approach carries inherent blind spots that motivate fusion.

\paragraph{Rainfall-only models.}
Rainfall data is the most widely used input for early flood warning.
Muñoz et al.\ \cite{munoz2021tomebamba} applied five ML classifiers,
including multi-layer perceptron (MLP) and random forest, to forecast three
flood-alert levels (No-alert, Pre-alert, Alert) in the Tomebamba Andean
catchment using only precipitation and discharge from rain gauges.
The best model (MLP) achieved an F1-macro of 0.82 for 1-hour lead time,
degrading sharply beyond 6~hours.  The authors explicitly acknowledged that
precipitation-runoff models reach an effectiveness ceiling without additional
variables such as soil moisture or spatial flood confirmation.
Duncan et al.\ \cite{duncan2013rapids} developed the RAPIDS urban drainage
early warning system using radar-derived precipitation as the sole input
to forecast manhole flooding in near real time, but noted that forecasting
capability is fundamentally bounded by the time of concentration of the
catchment.  Similarly, Hayder et al.\ \cite{hayder2023intelligent} built
an ES-LSTM and ANN framework using rainfall, wind speed, and temperature
from the Australian meteorological dataset; while classification accuracy
exceeded 90\%, the authors noted that the absence of water-level observations
makes flood onset detection unreliable when rainfall signals are ambiguous.

\paragraph{Discharge/water-level-only models.}
Sensor-only approaches directly capture hydrological response but miss
the \emph{causal} precipitation signal that provides early warning before
depth increases materialize.  Guo et al.\ \cite{guo2021iotlora} deployed
an IoT flood early detection system using LoRaWAN-connected gauging stations,
applying ANN, LSTM, and GRU models to forecast discharge; the system achieved
strong nowcasting performance but provided no early warning capability ahead
of the rainfall-to-runoff lag.  Cruz et al.\ \cite{cruz2018tencon} used a
multi-layer ANN with rain gauge and water-level sensor inputs, supplemented
by soil-moisture and anemometer data, to forecast flood levels in the
typhoon-prone Philippines; despite good fit on training data, the single-site
nature of the sensor fusion limits spatial generalisability.

\paragraph{Humidity, temperature, and pressure models.}
Several studies have forecasted precipitation itself, a proxy for flood
precursors, from ambient meteorological variables.  Sarangpure et al.\
\cite{sarangpure2022multilayer} trained a multi-layer ANN using temperature,
relative humidity, barometric pressure, wind speed, and a rain gauge to
forecast water levels in the Philippines, obtaining goodness-of-fit values
above 0.93.  While promising, such meteorological-only models conflate
atmospheric predictors with hydrological outcomes; they cannot confirm
actual inundation and are prone to false alarms during periods of high
humidity without flooding.

\paragraph{Satellite-only models.}
Satellite-derived flood extent provides valuable spatial context but suffers
from coarse temporal resolution and data gaps caused by cloud cover.
When used in isolation, satellite signals tend to confirm flooding
\emph{after} it has begun rather than providing actionable advance warning.
As shown in our ablation study, a satellite-only
LSTM achieves the highest accuracy (0.956) among unimodal baselines yet the
lowest macro F1 (0.351), because the model collapses to forecasting no-flood
at nearly all timesteps and exploits class imbalance rather than learning
genuine flood dynamics.

\paragraph{The myopia of single-modality sensing.}
Across these works, a consistent pattern emerges: any single sensing modality
is \emph{myopic}.  Rainfall sensors fire too early in the causal chain and
produce false alarms during high-intensity events that do not result in
flooding.  Water-level sensors fire too late, confirming a flood that has
already begun.  Temperature and atmospheric-pressure models lack the spatial
and hydrological specificity needed to distinguish flood conditions from
ordinary weather variation.  Satellite imagery provides spatial coverage but
cannot sustain the temporal resolution needed for real-time alerting.  A
sensor failure, whether from communication outage, fouling, or power loss, silences the
entire system.  These limitations are not incidental; they reflect the
fundamental information-theoretic constraint that no single view can capture
the full joint distribution of a multi-factor physical process.

\subsection{Multimodal Data Integration for Flood Detection}

Recognition of the myopia problem has motivated a growing strand of research
on data fusion for flood monitoring.

Mosavi et al.\ \cite{mosavi2018review} provide a comprehensive review of ML
models for flood forecasting, concluding that hybrid and ensemble methods that
combine meteorological and hydrological inputs consistently outperform
unimodal alternatives.  Darkwah et al.\ \cite{darkwah2024wcsna} developed
an ML-based flood forecasting system for the Window Cliffs State Natural
Area (the deployment site for the present work), combining LSTM, random forest,
and support vector regression with NWS rainfall forecasts and USGS gauge
data; this study directly informs the operational context and baseline
performance against which our multimodal framework is evaluated.

Beyond sensor fusion, recent work has explored fusing vision-based depth
estimation with numerical models.  Chang et al.\ \cite{chang2019hydroinfo}
built a regional flood inundation warning system that combines radar
precipitation forecasts with ANN-based runoff modelling;
Tan et al.\ \cite{tan2023algae} demonstrated that stationary RGB camera
images can be converted into quantitative environmental indices via CNN
pipelines, a principle directly analogous to our image-to-time-series
conversion of camera and satellite data.  Image-based flood monitoring
has also been explored through IoT and edge-computing lenses
\cite{9239693,guo2021iotlora}, and several works have shown that
infrared and visible-spectrum cameras can detect water surfaces under
diverse lighting conditions \cite{kruger2016bridge,kalyanapu2023lowcost}.

Remote sensing integration represents a complementary direction.
Portalés-Juliá et al.\ \cite{portalescjulia2023} and Munasinghe et al.\
\cite{munasinghe2023multisensor} show that combining optical and SAR
satellite observations increases flood-mapping accuracy and spatial
coverage, but the multi-day revisit cycles limit their utility for
real-time alerting.  Our framework addresses this by treating satellite
depth as a \emph{periodic supplementary} modality within a continuous
15-minute temporal pipeline anchored by camera and sensor streams.

Despite these advances, a gap remains: prior multimodal approaches
either (a) fuse only two or three modalities, (b) operate at coarse
spatial or temporal resolution, or (c) lack a reasoning layer to
enforce logical consistency of forecasted flood stages.  The work of
Helmrich et al.\ \cite{helmrich2021} on crowd-sourced monitoring
explicitly calls for multi-source integration to overcome data sparsity,
while Bales and Wagner \cite{bales2009uncertainty} identify inconsistent
stage transitions, a consequence of noisy or fragmented sensing, as
a principal source of uncertainty in flood inundation maps.

\subsection{LSTM Networks and Rule-Based Reasoning for Temporal Environmental Data}

Long Short-Term Memory networks \cite{hochreiter1997lstm} have emerged as
the dominant architecture for hydrological time-series modelling, owing to
their ability to learn long-range temporal dependencies through gated memory
cells.  Kratzert et al.\ \cite{kratzert2018lstm} demonstrated that an LSTM
trained on rainfall-runoff records outperforms traditional conceptual models
across hundreds of catchments, establishing the viability of purely
data-driven deep learning for operational hydrology.
Fordjour and Kalyanapu \cite{fordjour2024gru} applied a GRU network (a
simplified LSTM variant) to flood forecasting in Ashland City, Tennessee,
achieving strong lead times on a local basin closely related to our study area.

The integration of rule-based or symbolic reasoning with neural-network
forecasts has received attention in safety-critical domains where model
outputs must be operationally interpretable.  Flood stage forecasting is
one such domain: an alert system that oscillates between ``moderate'' and
``no-flood'' classifications at consecutive timesteps is operationally
useless regardless of average accuracy.  Teng et al.\ \cite{teng2017review}
identify temporal inconsistency as a principal failure mode in flood
inundation models; our minimum-duration reasoning layer directly addresses
this failure mode by suppressing isolated short-duration anomalies while
preserving abrupt multi-stage transitions characteristic of flash flooding.

\subsection{Summary: Gaps Addressed by the Present Work}

Table~\ref{tab:comparison} summarises prior flood detection and forecasting
systems in terms of sensing modalities, temporal resolution, spatial coverage,
and key limitations.  The proposed multimodal framework is the only system
that (a) jointly models four heterogeneous modalities (hydrometer, rainfall,
camera, and satellite) within a single temporal pipeline, (b) converts
visual observations into time-series depth signals enabling end-to-end
differentiable learning, and (c) augments neural forecasting with a
rule-based reasoning layer that stabilises stage transitions without
sacrificing flash-flood responsiveness.

\begin{table*}[t]
\caption{Comparison of flood detection and forecasting systems.
``Mod.'' = number of sensing modalities.  $\checkmark$ = supported;
$\times$ = not supported; $\sim$ = partial.}
\label{tab:comparison}
\centering
\small
\setlength{\tabcolsep}{4pt}
\begin{tabularx}{\textwidth}{lccccccX}
\toprule
System & Mod. & Rainfall Depth & Sensor Depth
  & Camera & Satellite & Reasoning Layer
  & Key Limitation \\
\midrule
Muñoz et al.\ \cite{munoz2021tomebamba}
  & 2 & $\checkmark$ & $\checkmark$ & $\times$ & $\times$ & $\times$
  & Accuracy degrades beyond 6-h lead time \\
Duncan et al.\ \cite{duncan2013rapids}
  & 1 & $\checkmark$ & $\times$ & $\times$ & $\times$ & $\times$
  & Limited to time of concentration \\
Hayder et al.\ \cite{hayder2023intelligent}
  & 3 & $\checkmark$ & $\times$ & $\times$ & $\times$ & $\times$
  & No water-depth confirmation \\
Guo et al.\ \cite{guo2021iotlora}
  & 2 & $\checkmark$ & $\checkmark$ & $\times$ & $\times$ & $\times$
  & No early warning; single modality at edge \\
Sarangpure et al.\ \cite{sarangpure2022multilayer}
  & 4 & $\checkmark$ & $\checkmark$ & $\times$ & $\times$ & $\times$
  & Atmospheric proxies; no visual confirmation \\
Darkwah et al.\ \cite{darkwah2024wcsna}
  & 2 & $\checkmark$ & $\checkmark$ & $\times$ & $\times$ & $\times$
  & No camera or satellite integration \\
Latham et al.\ \cite{latham2025aquacam}
  & 1 & $\times$ & $\times$ & $\checkmark$ & $\times$ & $\times$
  & Single-site, no forecasting \\
Chang et al.\ \cite{chang2019hydroinfo}
  & 2 & $\checkmark$ & $\sim$ & $\times$ & $\times$ & $\times$
  & Regional scale; no real-time camera \\
Munasinghe et al.\ \cite{munasinghe2023multisensor}
  & 2 & $\times$ & $\times$ & $\times$ & $\checkmark$ & $\times$
  & Low temporal resolution; cloud cover \\
\midrule
This work
  & 4 & $\checkmark$ & $\checkmark$ & $\checkmark$ & $\checkmark$
  & $\checkmark$
  & Simulation-derived training data (see Sec.~\ref{sec:limitations}) \\
\bottomrule
\end{tabularx}
\end{table*}
\section{Methodology}

\subsection{Problem Formulation}
We formulate flood stage forecasting as a multimodal time-series classification problem. At each timestep $t$, the system receives a synchronized tuple of observations from four sensing modalities:
\begin{equation}
\mathbf{x}_t = \left[ x_t^{\text{rain}},\ x_t^{\text{sensor}},\ x_t^{\text{camera}},\ x_t^{\text{sat}} \right] \in \mathbb{R}^4
\end{equation}
where $x_t^{\text{rain}}$ denotes rainfall depth, $x_t^{\text{sensor}}$ denotes hydrometer-measured water depth, $x_t^{\text{camera}}$ denotes camera-derived average inundation depth, and $x_t^{\text{sat}}$ denotes average inundation depth extracted from satellite imagery. All signals are sampled at 15-minute intervals and temporally aligned to a common timeline.

Given a lookback window of $L$ consecutive timesteps ending at time $t$, the input to the model is a multimodal sequence:
\begin{equation}
\mathbf{X}_t = \left[ \mathbf{x}_{t-L+1},\ \mathbf{x}_{t-L+2},\ \ldots,\ \mathbf{x}_t \right] \in \mathbb{R}^{L \times 4}
\end{equation}

Critically, the objective is to forecast the flood stage label $y_{t+h} \in \{\text{no\_flood},\ \text{low},\ \text{moderate},\ \text{severe}\}$ at a future timestep $t+h$, $h$ steps \emph{ahead} of the input window, using only observations up to and including $t$. The forecast horizon $h \geq 1$ is enforced strictly, so the model never has access to any observation at or after the timestep it is forecasting; this is what distinguishes genuine forecasting from same-timestep classification, in which the model would forecast $y_t$ from a window that already ends at $t$. Three horizons are evaluated: $h \in \{1, 4, 12\}$, corresponding to 15 minutes, 1 hour, and 3 hours ahead at 15-minute resolution. The four flood stages correspond to increasing levels of inundation severity, defined through threshold-based labeling applied to all four sensing modalities with majority voting, as described in Section~\ref{sec:labeling}.

As a lower bound on genuine forecasting skill, we additionally report a trivial persistence baseline that requires no training: $\hat{y}_{t+h} = y_t$, i.e., the stage is assumed unchanged from the most recent observation. A trained model that fails to exceed this baseline provides no forecasting value beyond simple persistence, and this comparison is reported alongside every result in Section~5.

The primary objectives of the system are: (1) accurately classify future flood stages across all severity levels, with particular emphasis on minority flood classes; (2) maintain stable and consistent stage forecasts that minimize spurious oscillations; and (3) provide modality-level explanations for each forecasted flood stage to support interpretable decision-making.

\subsection{Dataset Construction}

\subsubsection{Study Area}
\label{sec:study-area}

\begin{figure}[!htbp]
\centering
\includegraphics[width=\linewidth]{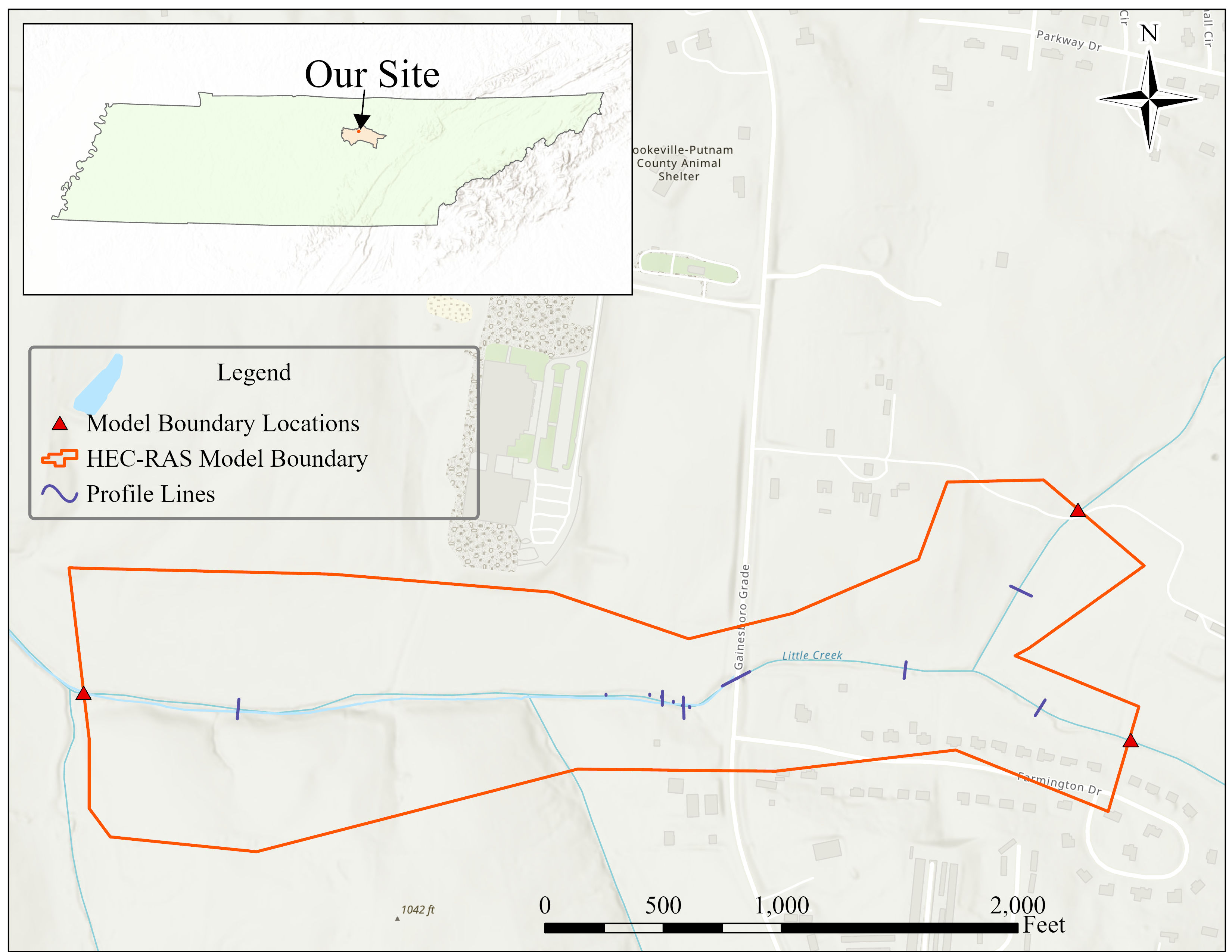}
\caption{Study area: Little Creek, West Blackburn Fork watershed, Putnam County, Tennessee.}
\label{fig:study-area}
\end{figure}

The hydraulic simulation underlying the sensor, camera, and satellite modalities (Sections~\ref{sec:study-area}--\ref{sec:satellite-depth}) is built on Little Creek, a tributary of the West Blackburn Fork in the West Blackburn Fork watershed, Cookeville, Putnam County, Tennessee. The HEC-RAS model uses an unstructured computational mesh of 23,245 cells (average face length 13 ft, average cell size 173 ft$^2$, ranging from 6 ft$^2$ to 2,461 ft$^2$). Fig.~\ref{fig:study-area} shows the model boundary, the three model boundary locations, and the profile-line transects used for camera-derived depth (Section~\ref{sec:camera-depth}). Rainfall forcing for the simulation is a separate NOAA gauge record, described next.

\subsubsection{Rainfall Depth (Precipitation)}

Rainfall depth data is sourced from NOAA QPCP (Quantitative Precipitation - Cooperative) records for the Livingston, Tennessee gauge, providing 15-minute accumulated rainfall depth totals used as forcing input to the HEC-RAS simulation of the Little Creek watershed (Section~\ref{sec:study-area}).

\subsubsection{Hydrometer Data (Sensor Depth)}

Simulated hydrometer depth data is obtained from HEC-RAS hydraulic simulation outputs for the Little Creek study area (Section~\ref{sec:study-area}), driven by the Livingston NOAA rainfall record as input forcing. The simulation produces water surface elevation (WSE) and ground elevation values for a spatial grid of cells covering the study domain. For each simulation cell, the water depth at timestep $t$ is computed as:

\begin{equation}
d_t^{\text{sensor}} = \text{WSE}_t - \text{GroundElev}
\end{equation}

\subsubsection{Camera-Derived Depth (Profile Line)}
\label{sec:camera-depth}

In the current simulation-based setup, camera-derived water depth is approximated using HEC-RAS outputs, which correspond to a cross-sectional transect representative of inundation in the Little Creek study area (Section~\ref{sec:study-area}, Fig.~\ref{fig:study-area}). Rather than processing actual camera imagery, this approach uses hydraulic simulation outputs as a proxy for what a deployed camera system would observe. The profile line output provides water surface elevation, ground elevation, and inundation depth at multiple spatial points along the transect at each timestep. The average depth across all profile line points is computed at each timestep to produce a single camera-derived depth time series.
\subsubsection{Satellite-Derived Depth}
\label{sec:satellite-depth}

Satellite-derived inundation depth is extracted from a series of simulated flood depth raster images, generated from HEC-RAS simulation outputs for the study domain. Each image represents the spatial distribution of estimated flood depth at a specific simulation timestep, encoded as color values mapped to depth ranges. A pixel-level extraction pipeline reads each image, maps pixel colors to corresponding depth values, and computes the average depth across all inundated pixels to produce a single scalar value per image. \\

Missing values across all four modalities (most commonly satellite depth, due to its lower temporal resolution) are imputed using forward-fill followed by backward-fill, propagating the most recent available observation forward in time.

\subsection{Flood Stage Labeling}
\label{sec:labeling}

Ground truth flood stage labels are generated using a threshold-based majority voting scheme applied independently to each of the four sensing modalities. For each modality, per-timestep depth or precipitation values are mapped to one of four flood stages (no\_flood, low, moderate, or severe) using the fixed thresholds in Table~\ref{tab:label-thresholds}, calibrated to the observed value ranges in the training data. For each timestep, the four per-modality stage labels are aggregated by majority vote; ties are resolved toward the lower-severity label, matching the exact tie-breaking behavior of the original labeling script this scheme reproduces. This threshold-and-vote scheme was reimplemented independently from that prior script and validated to reproduce the existing labels exactly (zero mismatches across all training-period rows) before being applied to backfill labels onto the test period, which originally had none.

\begin{table}[!htbp]
\centering
\small
\begin{tabular}{lccc}
\hline
Modality & low & moderate & severe \\
\hline
Sensor depth (ft)       & $\geq 0.5$ & $\geq 1.5$ & $\geq 2.5$ \\
Camera depth (ft)       & $\geq 0.1$ & $\geq 0.5$ & $\geq 1.0$ \\
Rainfall depth (in)     & $\geq 0.1$ & $\geq 0.3$ & $\geq 0.5$ \\
Satellite depth (ft)    & $\geq 0.7$ & $\geq 1.2$ & $\geq 1.8$ \\
\hline
\end{tabular}
\caption{Per-Modality Flood Stage Thresholds}
\label{tab:label-thresholds}
\end{table}

\begin{figure}[!htbp]
\centering
\includegraphics[width=\linewidth]{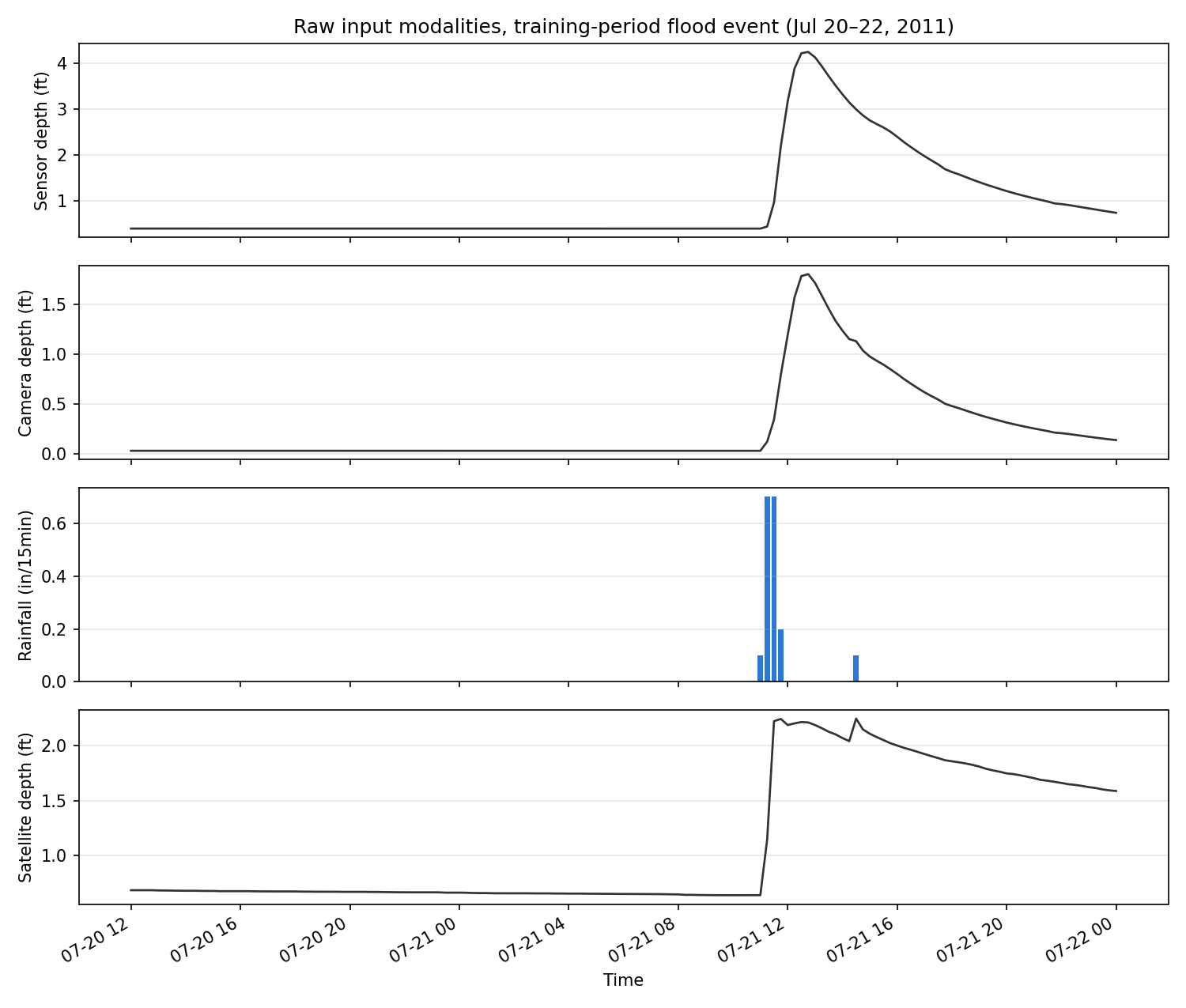}
\caption{Raw input modalities across one training-period flood event.}
\label{fig:raw-data-sample}
\end{figure}

Fig.~\ref{fig:raw-data-sample} shows all four raw input modalities across a single representative flood event from the training period, illustrating the relative timing and magnitude of each signal's response as the event unfolds.

It is important to acknowledge that this labeling scheme is synthetic: ground truth labels are derived from the same simulation and sensor data used as model inputs, rather than from independent field observations or gauge records. This is a known limitation of the current simulation-based experimental setup and is discussed further in Section~\ref{sec:limitations}.

\subsection{Data Preprocessing}

Prior to model training, all four feature channels are standardized using z-score normalization:

\begin{equation}
\tilde{x}^{(k)} = \frac{x^{(k)} - \mu^{(k)}}{\sigma^{(k)}}
\end{equation}

where $\mu^{(k)}$ and $\sigma^{(k)}$ are the mean and standard deviation of feature $k$, computed only on the portion of the training period preceding the validation cutoff (Section~\ref{sec:split}) so that no validation- or test-period statistics leak into scaling. Flood stage labels are encoded as integers, with the ordinal relationship between stages fixed explicitly rather than inferred alphabetically: no\_flood = 0, low = 1, moderate = 2, severe = 3.

\subsection{Sequence Construction}

The forecasting problem is formulated using a sliding window approach. For each valid anchor timestep $t$, a fixed-length input sequence of $L$ consecutive multimodal observations ending at $t$ is constructed and paired with the label at $t+h$:
\begin{equation}
\mathbf{X}_t = \left[ \mathbf{x}_{t-L+1},\ \ldots,\ \mathbf{x}_t \right] \in \mathbb{R}^{L \times 4} \ \longrightarrow\ y_{t+h}
\end{equation}
The primary lookback window is $L = 12$ timesteps (3 hours), motivated by the typical lag between rainfall onset and peak water level response. Section~\ref{sec:window-sensitivity} additionally evaluates $L \in \{1, 2, 4, 12, 24, 40\}$ (15 minutes, 30 minutes, 1, 3, 6, and 10 hours) to test whether a shorter or longer historical context improves forecasts.

\subsection{Train / Validation / Test Split}
\label{sec:split}

The model is trained on one simulated period and evaluated on a separate simulated period from a different year, with no temporal overlap between the two. Within the training period, the final 15\% (by chronological order, not a random split) is held out as a validation set, used only for hyperparameter selection and early stopping; the preceding 85\% is the fit set used for gradient updates. The test period is never used for any tuning decision. This chronological split, rather than a random one, avoids leaking autocorrelated future information into validation, and mirrors how the model would be used operationally: tuned only on the past, evaluated on genuinely held-out data.

\subsection{Class Imbalance Handling}
\label{sec:class-imbalance}

The training dataset is severely imbalanced, with no-flood conditions comprising 84.0\% of all timesteps. Naive training on this imbalanced distribution would cause the model to collapse toward forecasting no-flood at all times, achieving high accuracy while completely failing to detect genuine flood events. To address this, class-balanced weights are applied to the cross-entropy loss during training:

\begin{equation}
w_c = \frac{N}{C \cdot N_c}
\end{equation}

where $N$ is the total number of training samples, $C$ is the number of classes, and $N_c$ is the number of samples in class $c$. This formulation assigns higher loss weight to underrepresented classes, penalizing the model more severely for misclassifying rare flood events.

\subsection{Model Architecture}

\subsubsection{Architecture Overview}

\begin{figure}[h]
  \centering
  \includegraphics[width=\linewidth]{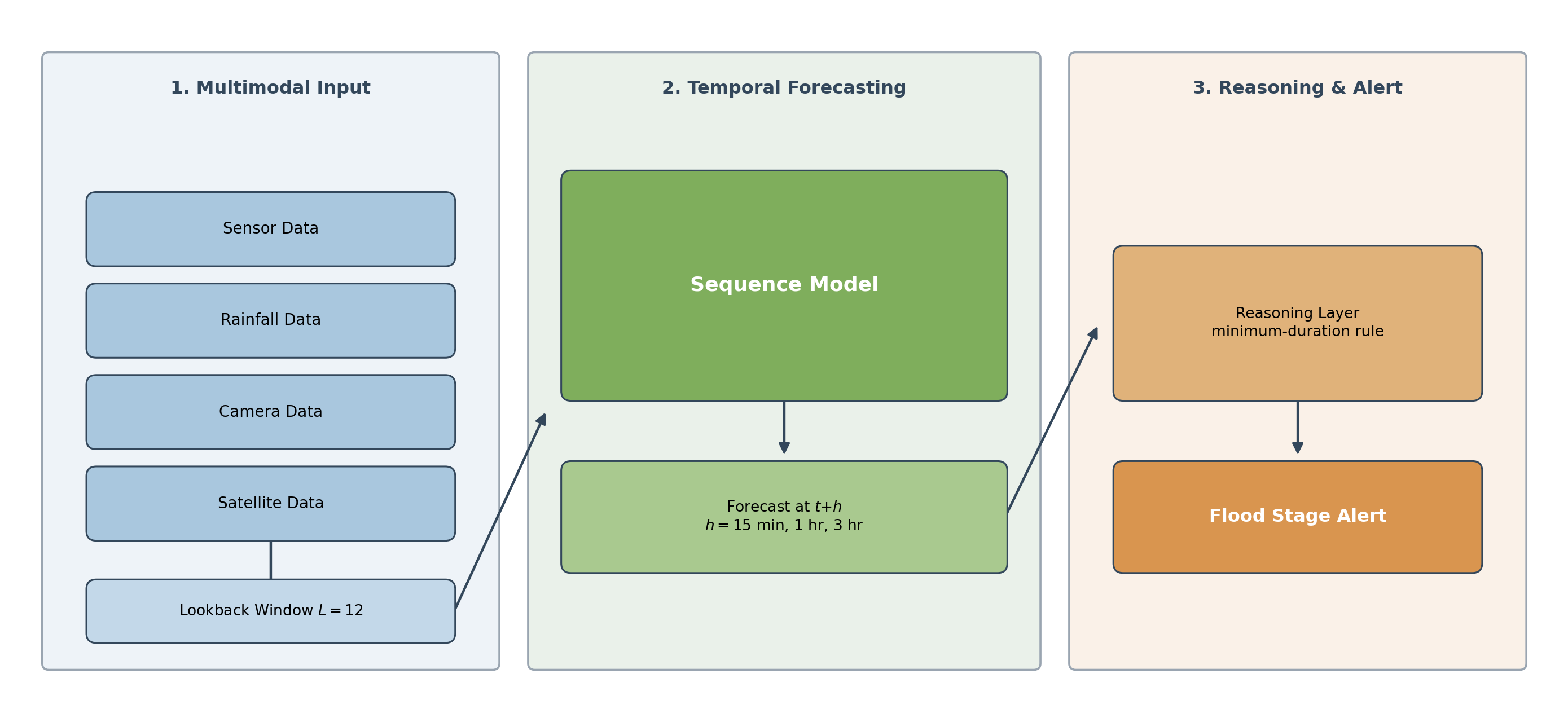}
  \caption{Overview of the multimodal flood detection architecture.}
  \label{fig:architecture}
\end{figure}

Fig.~\ref{fig:architecture} presents an overview of the system architecture. The pipeline consists of three main stages: multimodal input fusion, temporal sequence modeling, and rule-based reasoning. In the first stage, rainfall, sensor depth, and camera depth are concatenated as direct numerical inputs, while satellite depth is first reduced from raster imagery to a scalar via the fixed pixel-to-depth colorbar lookup procedure described in Section~\ref{sec:satellite-depth}. This is a deterministic table lookup and spatial averaging, not a learned (e.g., CNN-based) image encoder, and no spatial feature learning occurs at this stage. In the second stage, the fused multimodal sequence is passed to one of six candidate sequence models (Section~\ref{sec:architectures}) that learns joint temporal dependencies and produces a flood stage forecast. In the third stage, the raw forecasts pass through a rule-based reasoning layer that suppresses isolated short-duration false alarms and provides modality-level attribution for each alert.

\subsubsection{Fusion Strategy}
\label{sec:fusion}

Two multimodal fusion strategies are compared for every architecture. Early fusion concatenates all four modalities into a single feature vector at each timestep before the sequence model, allowing the model to learn cross-modal temporal dependencies directly from the raw multimodal input. Late fusion instead assigns a separate, smaller encoder to each modality; the four per-modality final hidden states are concatenated only immediately before the classification head, so cross-modal interaction is limited to the last layer rather than learned throughout the sequence. For the two tree-based architectures (Random Forest, XGBoost), which produce no intermediate hidden representation to fuse, the late-fusion analogue is an ensemble: four already-trained single-modality models are reused as-is, and their forecasted class probabilities are averaged. This requires no additional training for the tree-based late-fusion variants.

\subsubsection{Candidate Architectures}
\label{sec:architectures}

Six architectures are evaluated under identical data, training, and evaluation conditions, differing only in the sequence encoder (or, for the tree models, the input representation):

LSTM. A two-layer stacked Long Short-Term Memory network. LSTMs capture long-range temporal dependencies through gated memory cells, selectively retaining or discarding information over multiple timesteps. The final hidden state of the second layer is mapped to the four flood-stage logits by a linear layer:
\begin{equation}
\hat{y}_{t+h} = \mathbf{W} \cdot \mathbf{h}_t^{(2)} + \mathbf{b}
\end{equation}

GRU. Structurally identical to the LSTM encoder, with the LSTM cell replaced by a Gated Recurrent Unit, which uses fewer gates and no separate cell state.

TCN. A Temporal Convolutional Network: a stack of causal, dilated 1D convolutional blocks (kernel size 3, dilations $\{1, 2, 4\}$) with residual connections. The receptive field of the stack exceeds the lookback window, so the final layer's output already integrates information from every input timestep without recurrence.

Transformer. A linear input projection followed by sinusoidal positional encoding and a 2-layer Transformer encoder with multi-head self-attention. The representation at the final timestep is used for classification, for parity with the recurrent models' "final hidden state" semantics.

Random Forest and XGBoost. Rather than processing the sequence recurrently, the $L \times 4$ input window is flattened into a single feature vector and passed to a class-weighted Random Forest or XGBoost classifier. These serve as strong non-sequential baselines, testing whether explicit temporal modeling is necessary at all for this task.

\subsection{Hyperparameter Tuning}
\label{sec:tuning}

Each architecture's hyperparameters are tuned once, not once per configuration or horizon, via a 9-point grid search evaluated on the validation split (Section~\ref{sec:split}), using the full multimodal input at $h=1$. For the four neural architectures, the grid spans hidden size $\in \{32, 64, 128\}$ and dropout $\in \{0.2, 0.3, 0.4\}$; for Random Forest, $n_{\text{estimators}} \in \{100, 200, 400\}$ and max depth $\in \{\text{None}, 10, 20\}$; for XGBoost, $n_{\text{estimators}} \in \{100, 200, 400\}$ and max depth $\in \{3, 6, 9\}$, with learning rate, subsample, and column-subsample held fixed. The winning configuration for each architecture is then held fixed across all other modality configurations, horizons, and seeds for that architecture. Late-fusion sequence-encoder variants (LSTM, GRU, TCN, Transformer) reuse their early-fusion counterpart's tuned hyperparameters directly, since it is the same underlying cell fused differently, rather than being tuned a second time. This grid search doubles as sensitivity-analysis evidence that performance does not hinge on one arbitrarily lucky hyperparameter setting; the full grid is reported in Section~4.

\subsection{Training Procedure}

The model is trained using the Adam optimizer with default momentum parameters ($\beta_1 = 0.9$, $\beta_2 = 0.999$). The weighted cross-entropy loss is used as the training objective:

\begin{equation}
\mathcal{L} = -\sum_{c=1}^{C} w_c \cdot y_c \cdot \log(\hat{p}_c)
\end{equation}

where $y_c$ is the one-hot encoded true label, $\hat{p}_c$ is the forecasted probability for class $c$, and $w_c$ is the class weight defined in Section~\ref{sec:class-imbalance}. For the four neural architectures, training proceeds with early stopping on validation macro-F1 (patience of 8 epochs, up to a maximum of 50 epochs), and the model weights from the best validation epoch are restored before final test evaluation. Random Forest and XGBoost are trained to convergence directly (no epoch-based stopping applies to tree ensembles). All hyperparameter values are described in the Experimental Setup section. The scaler and label mapping fitted on the training set are saved alongside the model weights to ensure consistent preprocessing during inference on new data.

\subsection{Statistical Evaluation Protocol}
\label{sec:stats}

Every (architecture, fusion strategy, modality configuration, horizon) cell is trained and evaluated across 5 random seeds, and results are reported as mean $\pm$ standard deviation. Comparisons between two conditions (e.g., two architectures, or early vs. late fusion) use a paired evaluation matched by seed: both a paired $t$-test and a Wilcoxon signed-rank test are reported via scipy.stats. With only 5 paired seeds, the smallest possible two-sided Wilcoxon $p$-value is 0.0625, so even a unanimous 5-of-5 result cannot cross the conventional 0.05 threshold on that test alone at this sample size; the $t$-test is therefore treated as the more informative measure at this sample size, with the Wilcoxon result reported alongside as a standard non-parametric check rather than a contradicting result.

\subsection{Robustness Evaluation}
\label{sec:robustness}

To test whether the model's performance is robust to degraded sensor data, and specifically whether any observed tolerance to missing data reflects genuine robustness or simply reduced reliance on a given modality, two corruption scenarios are evaluated on the test set only (training data and the fitted scaler are never touched):

Missing. An additional 30\% of the currently-observed values in a given modality column are randomly set to missing, seeded for reproducibility, before the existing forward-fill/back-fill imputation runs as usual.

Wrong values. 30\% of currently-observed values in a given modality are instead overwritten with an implausible reading: the original value plus Gaussian noise scaled to 4$\times$ the column's own standard deviation, clipped at zero. Because the value is not missing, forward-fill cannot repair it, so the model receives an actively incorrect reading rather than a gap.

Each corruption type is applied independently to each of the four modalities in turn (the remaining three modalities stay clean), giving 8 scenarios. All 7 architecture variants (LSTM, LSTM late-fusion, GRU, TCN, Transformer, Random Forest, XGBoost) are evaluated using their already-trained clean-data multimodal checkpoints, with no retraining; clean-condition scores are reused directly from the main classification results, and significance is assessed with the paired protocol of Section~\ref{sec:stats}, matched by seed, for every (architecture, modality, corruption type, horizon) combination.

\subsection{Lookback-Window Sensitivity}
\label{sec:window-sensitivity}

To test whether a shorter or longer input history improves forecasts, six lookback windows ($L \in \{1, 2, 4, 12, 24, 40\}$, i.e., 15 minutes, 30 minutes, 1, 3, 6, and 10 hours) are evaluated against five forecast horizons ($h \in \{1, 4, 8, 12, 20\}$, i.e., 15 minutes, 1, 2, 3, and 5 hours), on the multimodal configuration for all six architectures. Hyperparameters are reused from Section~\ref{sec:tuning} rather than re-tuned per window size, consistent with the tuning protocol applied throughout.

\subsection{Reasoning Layer}

\subsubsection{Motivation}

Raw forecasts on temporal flood data can exhibit spurious oscillations, where the forecasted flood stage alternates rapidly between adjacent classes over consecutive timesteps. These oscillations arise from borderline classification decisions near decision boundaries, where small perturbations in the input sequence cause the model to alternate between two nearby classes. In a flood early warning context, such oscillations are operationally disruptive: an alert system that rapidly toggles between moderate and no-flood conditions is unreliable and erodes user trust, even if the individual forecasts are technically accurate on average.

A naive solution would be to enforce strict stage transition constraints, such as limiting the maximum stage change per timestep to one level. However, this approach is inappropriate for areas susceptible to flash flooding, a phenomenon characterized by abrupt, rapid onset of severe flooding with minimal warning time. Prohibiting multi-stage jumps would cause the system to systematically underestimate the severity of flash flood events, which are precisely the situations where rapid and accurate alerting is most critical.

\subsubsection{Minimum Duration Constraint}

We implement a minimum duration reasoning layer that suppresses isolated short-duration flood forecasts while preserving abrupt stage transitions. Formally, for a sequence of forecasted labels $\hat{y}_1, \hat{y}_2, \ldots, \hat{y}_T$, a forecasted flood stage at timestep $t$ is suppressed if it persists for fewer than $\tau = 3$ consecutive timesteps (corresponding to 45 minutes at 15-minute resolution). Suppressed forecasts are replaced with the surrounding baseline stage, which is no-flood in the majority of cases. This is a fixed, rule-based post-processing step applied to the primary LSTM multimodal forecasts; it uses no training data and learns no parameters of its own. Table~\ref{alg:reasoning} gives the complete procedure: a single left-to-right pass over the raw forecast sequence that identifies each maximal run of an identical non-no-flood stage, and, if that run is shorter than $\tau$, overwrites it with the reasoned value immediately preceding it (so a suppressed run merges into whichever stage came before it (possibly itself already reasoned), rather than being re-examined). No-flood runs are never suppressed, since there is no shorter stage to fall back to.

\begin{table}[!htbp]
\centering
\small
\begin{tabular}{l}
\hline
Algorithm 1: Minimum-Duration Reasoning Layer \\
\hline
Input: raw stage sequence $\hat{y}_1,\ldots,\hat{y}_T$; minimum duration $\tau$ \\
Output: reasoned stage sequence $y'_1,\ldots,y'_T$ \\[2pt]
\texttt{1:}\ \ $y' \leftarrow \hat{y}$ \quad (copy of the raw forecasts) \\
\texttt{2:}\ \ $i \leftarrow 1$ \\
\texttt{3:}\ \ while $i \leq T$ do \\
\texttt{4:}\ \ \quad $s \leftarrow y'_i$ \\
\texttt{5:}\ \ \quad if $s =$ no-flood then $i \leftarrow i{+}1$; continue \\
\texttt{6:}\ \ \quad $j \leftarrow i$ \\
\texttt{7:}\ \ \quad while $j \leq T$ and $y'_j = s$ do $j \leftarrow j{+}1$ \\
\texttt{8:}\ \ \quad \quad \emph{// $[i,j)$ is the maximal run of stage $s$ from $i$} \\
\texttt{9:}\ \ \quad if $j - i < \tau$ then \\
\texttt{10:} \quad\ \ $y'_i,\ldots,y'_{j-1} \leftarrow y'_{i-1}$ (or no-flood if $i=1$) \\
\texttt{11:} \quad $i \leftarrow j$ \\
\texttt{12:} return $y'$ \\
\hline
\end{tabular}
\caption{Minimum-Duration Reasoning Layer}
\label{alg:reasoning}
\end{table}

This constraint is motivated by the physical dynamics of flood events: genuine flooding, even in flash flood scenarios, involves sustained inundation that persists for at least several consecutive observation intervals. A single isolated timestep of forecasted moderate flooding surrounded by no-flood forecasts is far more likely to represent a classification noise artifact than a genuine flood event of sub-15-minute duration.

The minimum duration threshold of $\tau = 3$ is selected to balance two competing objectives: suppressing short-duration noise while preserving short but genuine flood onset sequences. A threshold of 45 minutes is consistent with the minimum meaningful flood duration in the HEC-RAS simulation outputs for the study area, where even brief rainfall events produce sustained depth responses of at least 2--3 timesteps. Critically, the reasoning layer places no constraint on the magnitude of stage transitions, allowing the forecasted stage to jump from no-flood directly to severe within a single timestep when supported by the model's confidence scores. This design explicitly accommodates flash flood dynamics.

To be explicit about how this rule is chosen and validated, since it is easy to conflate a fixed post-processing rule with a learned or tuned one: $\tau = 3$ is set once, from the physical reasoning above, before any evaluation is run; it is never fit, searched over, or adjusted based on validation or test performance, and it is applied identically to every architecture and every seed. What \emph{is} empirically evaluated is the effect of applying this fixed rule to the model's raw forecasts, reported in Section~\ref{sec:ablation-study} (Table~\ref{tab:reasoning}): both the oscillation count and the macro F1 score are measured on the same forecasts before and after the rule is applied, so its impact on stability and accuracy is demonstrated on held-out data rather than assumed from the design rationale alone.

\subsubsection{Explainability}
In addition to stage smoothing, the reasoning layer provides natural-language explanations for each forecasted flood stage to support interpretable decision-making. Rather than reporting raw sensor values, the system describes the observed trend across the preceding lookback window in plain language, identifying which modalities indicate rising conditions and characterizing the overall flood onset pattern. This form of explanation is intended to be interpretable by emergency managers without requiring technical knowledge of the underlying model.

\section{Experimental Setup}

\subsection{Train/Validation/Test Split}

The model is trained on one simulated period and evaluated on a separate simulated period from a different year, with no temporal overlap between the two splits. This year-to-year temporal split is intentionally more rigorous than a random or within-period split, as it requires the model to generalize across different hydrological events from a different year rather than interpolating within a single continuous simulation. Within the training period, the final 15\% (chronologically, not randomly) is further held out as a validation set for hyperparameter tuning and early stopping (Section~\ref{sec:tuning-setup}); the test period is never used for any tuning decision. The training and test periods also differ substantially in their class distributions: the training set contains 12.4\% flood-stage samples (low, moderate, severe combined), while the test set contains only 5.1\%, reflecting the more quiescent nature of the test period outside of the two major flood events. This distributional shift provides a realistic assessment of model performance under conditions representative of operational deployment, where flood events are rare relative to normal flow conditions.

\subsection{Baseline and Ablation Configurations}

All six candidate architectures (LSTM, GRU, TCN, Transformer, Random Forest, XGBoost; Section~\ref{sec:architectures}) are trained and evaluated under an identical set of nine modality configurations, to isolate the effect of architecture choice from the effect of input composition. The nine configurations are: the full multimodal input; four unimodal baselines, each using a single modality (sensor-only, camera-only, rain-only, satellite-only) with input size 1; and four leave-one-out ablations, each omitting a single modality while retaining the remaining three (without-sensor, without-camera, without-rain, without-satellite) with input size 3. A larger performance drop when a modality is removed indicates greater dependence on that modality, providing a direct measure of each sensing source's contribution to the fused representation. Each of the four late-fusion sequence-encoder variants (LSTM, GRU, TCN, Transformer) is evaluated on a 5-configuration subset (multimodal plus the four ablations), since a single modality has nothing to fuse and its late-fusion result is identical to early fusion's. The two tree-based late-fusion variants are evaluated on the same 5-configuration subset, reusing already-trained single-modality models as described in Section~\ref{sec:fusion}.

\subsection{Hyperparameters}
\label{sec:tuning-setup}

Each architecture's hyperparameters are tuned once via the grid search described in Section~\ref{sec:tuning}, on the validation split, using the full multimodal input at $h=1$. The winning configuration per architecture, held fixed across all other configurations, horizons, and seeds, is summarized in Table~\ref{tab:hyperparams-neural} (neural architectures) and Table~\ref{tab:hyperparams-tree} (tree-based architectures); the TCN's kernel size is fixed at 3 and the Transformer's attention head count is fixed at 4, neither of which was part of the tuning grid. Late-fusion sequence-encoder variants reuse their early-fusion counterpart's values directly.

\begin{table}[!htbp]
\centering
\footnotesize
\setlength{\tabcolsep}{3pt}
\begin{tabular}{lcccc}
\hline
Arch. & Hidden & Depth & Drop. & LR \\
\hline
LSTM         & 32  & 2 & 0.2 & $10^{-3}$ \\
GRU          & 32  & 2 & 0.4 & $10^{-3}$ \\
TCN          & 64  & 3 & 0.2 & $10^{-3}$ \\
Transformer  & 128 & 2 & 0.4 & $10^{-3}$ \\
\hline
\end{tabular}
\caption{Tuned Hyperparameters, Neural Architectures}
\label{tab:hyperparams-neural}
\end{table}

\begin{table}[!htbp]
\centering
\small
\begin{tabular}{lccccc}
\hline
Arch. & $n_{\text{est}}$ & Depth & LR & Sub. & Col. \\
\hline
RF      & 200 & None & -- & -- & -- \\
XGBoost & 100 & 9    & 0.1 & 0.8 & 0.8 \\
\hline
\end{tabular}
\caption{Tuned Hyperparameters, Tree-Based Architectures}
\label{tab:hyperparams-tree}
\end{table}

In addition to the tuned values above, the following are held fixed across all architectures and experiments, summarized in Table~\ref{tab:hyperparams-fixed}.

\begin{table}[!htbp]
\centering
\small
\begin{tabular}{ll}
\hline
Hyperparameter & Value \\
\hline
Primary lookback ($L$)  & 12 steps (3 hr) \\
Optimizer (neural)      & Adam \\
Batch size              & 64 \\
Max epochs (neural)     & 50, patience 8 \\
Loss function           & Class-weighted CE \\
Seeds per cell          & 5 \\
Reasoning duration ($\tau$) & 3 steps (45 min) \\
\hline
\end{tabular}
\caption{Fixed Hyperparameters, All Experiments}
\label{tab:hyperparams-fixed}
\end{table}

\subsection{Robustness Experiment Setup}

Following the protocol in Section~\ref{sec:robustness}, robustness is evaluated on the test set only, at 30\% corruption severity (of currently-observed values), applied independently to each of the four modalities under each of the two corruption types (missing, wrong values), giving 8 scenarios. All 7 architecture variants are evaluated at all 3 horizons across 5 seeds, using already-trained clean-data checkpoints with no retraining, for a total of $7 \times 4 \times 2 \times 3 \times 5 = 840$ evaluations.

\subsection{Window/Horizon Sweep Setup}

Following the protocol in Section~\ref{sec:window-sensitivity}, six lookback windows (15 min, 30 min, 1, 3, 6, 10 hours) are crossed with five forecast horizons (15 min, 1, 2, 3, 5 hours) on all six architectures, multimodal configuration only, across 5 seeds, for a total of $6 \times 5 \times 6 \times 5 = 900$ training runs.

\subsection{Statistical Testing}

All comparisons follow the paired protocol described in Section~\ref{sec:stats}: a paired $t$-test and Wilcoxon signed-rank test, matched by seed. Four sets of comparisons are reported in Section~5: (1) multimodal LSTM vs. its own best unimodal baseline, per horizon; (2) early- vs. late-fusion, per architecture per horizon; (3) clean vs. corrupted test performance, per architecture per modality per corruption type per horizon; and (4) the primary 3-hour lookback window against both the shortest (15 min) and longest (10hr) windows evaluated, per architecture per horizon.

\subsection{Evaluation Metrics}

Given the severe class imbalance in both the training and test sets, macro-averaged F1-score is adopted as the primary evaluation metric. Macro F1 assigns equal weight to each flood stage class regardless of sample frequency, ensuring that performance on rare but operationally critical flood stages (moderate, severe) is not obscured by the dominant no-flood class. Per-class precision, recall, and F1 are reported to provide a detailed characterization of model behavior at each flood severity level.

Accuracy and weighted F1 are reported as secondary metrics for completeness, but are not used as the basis for model comparison due to their susceptibility to majority-class inflation under class imbalance. One-vs-rest ROC curves and AUC scores are computed for the multimodal model to assess the discriminative quality of the model's probabilistic outputs across all flood stages. Stage oscillation count, defined as the number of timestep-to-timestep stage transitions in the forecasted label sequence, is used as a stability metric to evaluate the effect of the reasoning layer on forecast consistency.

\section{Results}

The results below follow the same order the questions were raised in Section~\ref{sec:intro}. Sections~\ref{sec:flood-detection-performance}--\ref{sec:early-late-fusion} establish whether multimodal fusion outperforms any single sensor, and whether that holds across every forecast horizon. Sections~\ref{sec:ablation-study}--\ref{sec:window-sensitivity-results} examine what the reasoning layer, ablation, and lookback-window choices contribute once fusion itself is established. Section~\ref{sec:cross-modal-corr} and Section~\ref{sec:regression-threshold} stress-test two remaining concerns about the simulated setup directly rather than leaving them as caveats. Finally, Section~\ref{sec:real-world-validation} answers the second question from the introduction: whether any of this transfers to real field data, not only the simulation it was developed on.

\subsection{Training Convergence}

\begin{figure}[!htbp]
  \centering
  \includegraphics[width=\linewidth]{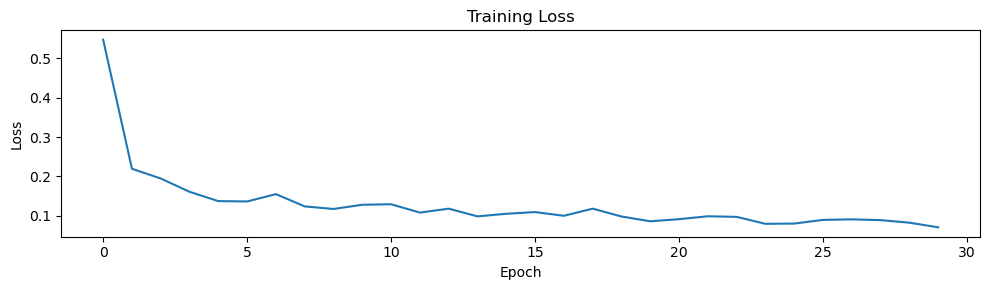}
  \caption{Training loss for the multimodal LSTM (representative seed).}
  \label{fig:loss}
\end{figure}

Fig.~\ref{fig:loss} presents the training loss curve for the multimodal LSTM (seed 0, $h=1$) over the course of training. The loss drops sharply in the first several epochs, reflecting rapid initial learning of dominant temporal patterns in the multimodal input, before stabilizing under early stopping. No oscillation or divergence is observed, indicating stable optimization under the class-weighted loss formulation.

\subsection{Flood Detection Performance}
\label{sec:flood-detection-performance}

\begin{figure}[!htbp]
  \centering
  \includegraphics[width=\linewidth]{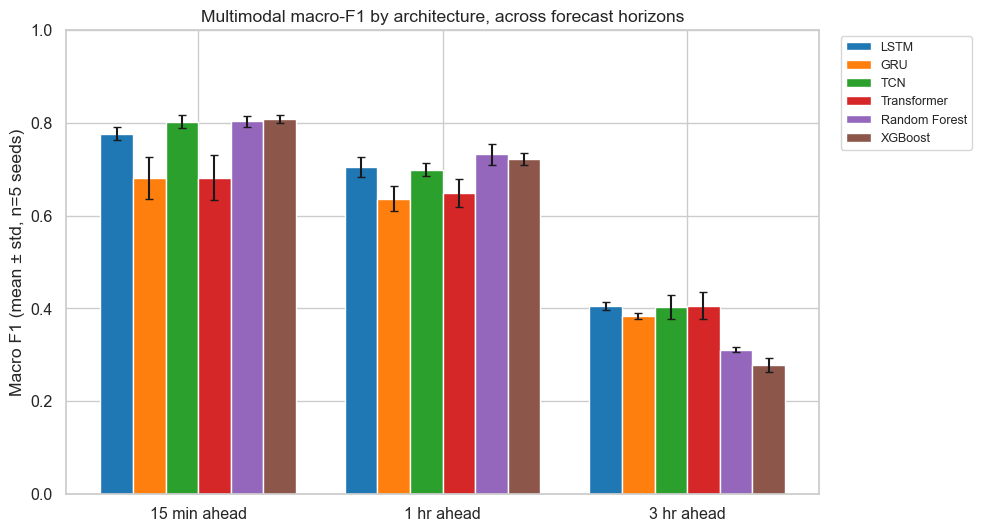}
  \caption{Macro F1 by architecture and forecast horizon.}
  \label{fig:arch_comparison}
\end{figure}

Table~\ref{tab:baseline} reports macro F1 (mean $\pm$ std across 5 seeds) for all six architectures on the full multimodal configuration, at all three forecast horizons, alongside the trivial persistence baseline (Section~3.1) at each horizon. Fig.~\ref{fig:arch_comparison} presents the same results as a bar chart.

\begin{table}[!htbp]
\centering
\small
\begin{tabular}{lccc}
\hline
Model & 15 min & 1 hr & 3 hr \\
\hline
Trivial baseline & 0.855 & 0.629 & 0.309 \\
\hline
LSTM         & 0.777 $\pm$ .014 & 0.705 $\pm$ .022 & 0.404 $\pm$ .009 \\
GRU          & 0.681 $\pm$ .046 & 0.636 $\pm$ .027 & 0.383 $\pm$ .006 \\
TCN          & 0.803 $\pm$ .014 & 0.699 $\pm$ .014 & 0.403 $\pm$ .026 \\
Transformer  & 0.682 $\pm$ .048 & 0.648 $\pm$ .030 & 0.406 $\pm$ .029 \\
Random Forest & 0.803 $\pm$ .012 & 0.732 $\pm$ .023 & 0.311 $\pm$ .005 \\
XGBoost      & 0.808 $\pm$ .008 & 0.722 $\pm$ .013 & 0.278 $\pm$ .015 \\
\hline
\end{tabular}
\caption{Macro F1 by Architecture and Forecast Horizon (Multimodal, 5-Seed Mean)}
\label{tab:baseline}
\end{table}

At the 15-minute horizon, no architecture exceeds the trivial persistence baseline (0.855); flood stage rarely changes within this interval, making ``no change'' a genuinely difficult target to beat, and this is reported as an inherent property of the forecasting task rather than a model shortcoming. At the 1-hour and 3-hour horizons, every architecture exceeds the trivial baseline. No single architecture is best at every horizon: XGBoost and Random Forest lead at 15 minutes and 1 hour, while LSTM, TCN, and Transformer are statistically indistinguishable at the 3-hour horizon and both tree-based models fall behind the trivial baseline at 3 hours. This horizon-dependence is examined directly in the following two subsections.

\subsection{Early versus Late Fusion}
\label{sec:early-late-fusion}

\begin{figure}[!htbp]
  \centering
  \includegraphics[width=\linewidth]{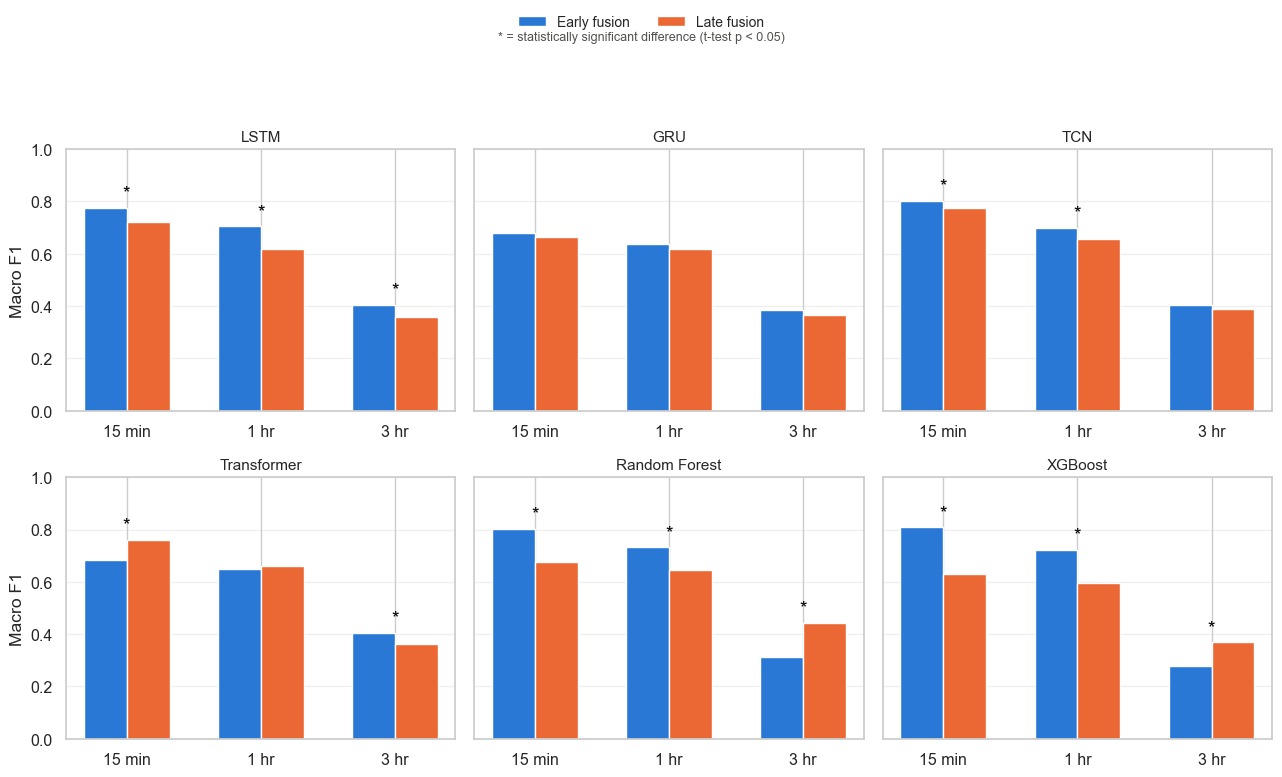}
  \caption{Early vs. late fusion by architecture and horizon (* = $p<0.05$).}
  \label{fig:fusion}
\end{figure}

For each of the six architectures, a late-fusion variant is trained and compared against its early-fusion counterpart (Section~\ref{sec:fusion}). Fig.~\ref{fig:fusion} presents macro F1 for both fusion strategies across all three horizons, with significance markers; the full paired-test results are summarized in the text below.

The result does not generalize uniformly across architectures, which is itself the key finding. LSTM shows a consistent, statistically significant advantage for early fusion at all three horizons. TCN significantly favors early fusion at 15 minutes and 1 hour, with no significant difference at 3 hours. GRU shows no statistically significant difference at any horizon. Transformer significantly favors \emph{late} fusion at 15 minutes, shows no difference at 1 hour, and favors early fusion at 3 hours. Random Forest and XGBoost both show a sharp, highly significant reversal: early fusion is substantially better at 15 minutes and 1 hour, but late fusion is substantially and significantly better at 3 hours. For both tree architectures, the late-fusion ensemble recovers performance that collapsed under early fusion at the longest horizon (Random Forest: 0.311 $\to$ 0.444; XGBoost: 0.278 $\to$ 0.369), moving from below or near the trivial persistence baseline (0.309) to clearly above it. The claim that early fusion outperforms late fusion is well-supported for LSTM specifically, but should not be generalized to all architectures: the appropriate fusion strategy, like the appropriate architecture, depends on the forecast horizon.

\subsection{Per-Class Classification Report}

Table~\ref{tab:multimodal_report} presents the per-class classification report for the primary multimodal LSTM model (representative seed, $h=1$).

\begin{table}[!htbp]
\centering
\begin{tabular}{lcccc}
\hline
Class & Precision & Recall & F1 & Support \\
\hline
no\_flood  & 1.00 & 0.97 & 0.98 & 2632 \\
low        & 0.58 & 0.74 & 0.65 & 87   \\
moderate   & 0.56 & 0.97 & 0.71 & 32   \\
severe     & 0.58 & 1.00 & 0.73 & 22   \\
\hline
macro avg & 0.68 & 0.92 & 0.77 & 2773 \\
\hline
\end{tabular}
\caption{Per-Class Report, Multimodal LSTM ($h=1$, Representative Seed)}
\label{tab:multimodal_report}
\end{table}

The model achieves high recall on the two most operationally critical classes: severe (1.00) and moderate (0.97). This means the model correctly identifies all severe flood events and 97\% of moderate events in the unseen test set, which is the most important property for a flood early warning system. Missing a severe flood event carries far greater consequences than generating a false alarm, and the model's design explicitly prioritizes this behavior through class-weighted training. This result is specific to LSTM at the shortest (15-minute) horizon, however; Section~\ref{sec:per-class-all} reports the same breakdown for all six architectures across all three horizons and finds it does not generalize uniformly.

The lower precision on moderate (0.56) and severe (0.58) indicates that some forecasts are false positives. However, these false positives predominantly occur as conservative over-forecasts, meaning the model forecasts a higher stage when the true label is one stage lower. Such adjacent-class misclassifications represent graceful degradation rather than catastrophic failure: the system issues a slightly elevated alert rather than missing the event entirely.

\begin{figure}[!htbp]
  \centering
  \includegraphics[width=\linewidth]{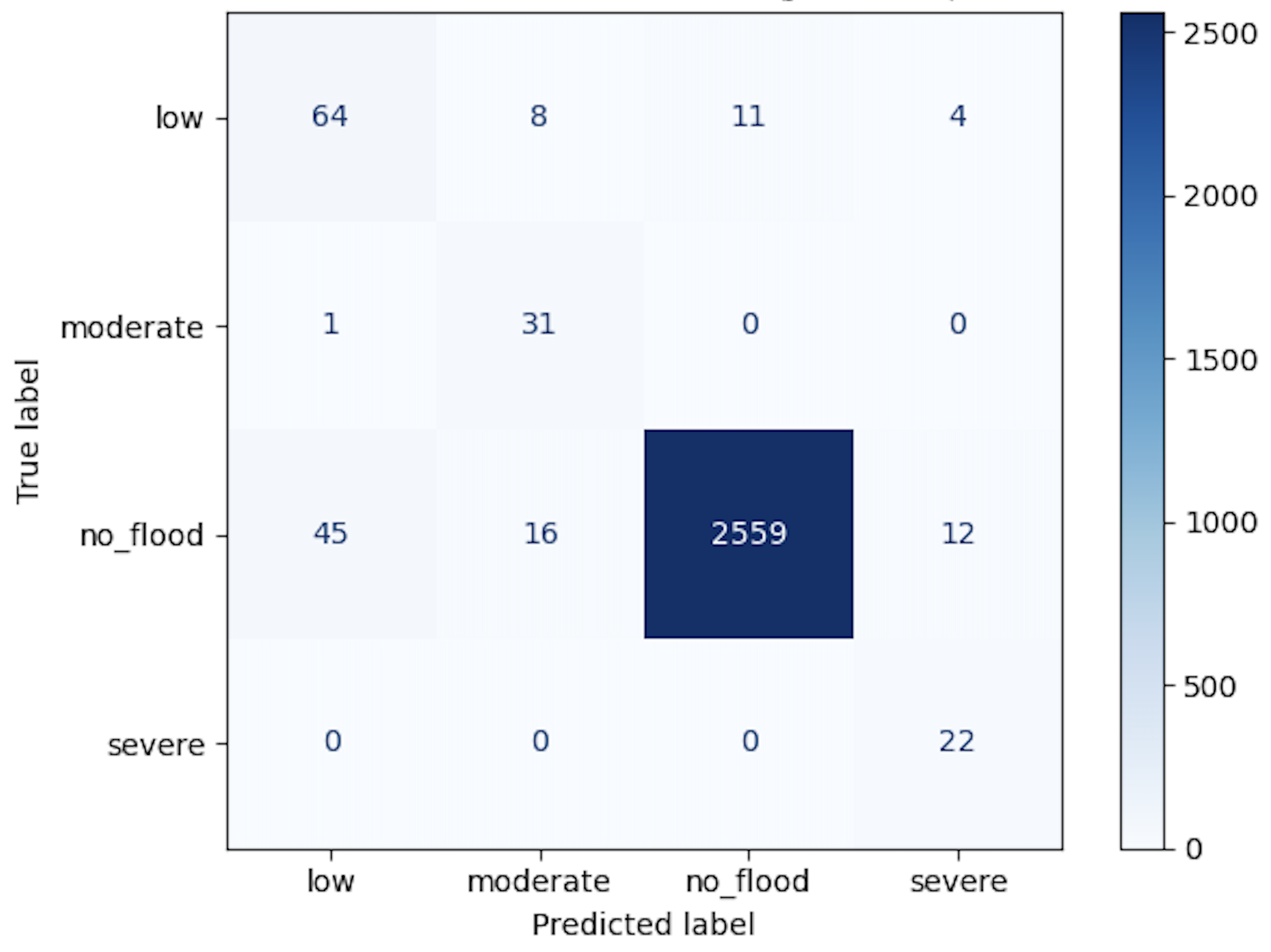}
  \caption{Confusion matrix, multimodal LSTM (representative seed, 15-min horizon).}
  \label{fig:cm_test}
\end{figure}

Fig.~\ref{fig:cm_test} shows the corresponding confusion matrix. Most misclassifications occur between adjacent flood stages, confirming the pattern of conservative over-forecast rather than catastrophic missed detections.

\subsection{ROC Analysis}

\begin{figure}[!htbp]
  \centering
  \includegraphics[width=\linewidth]{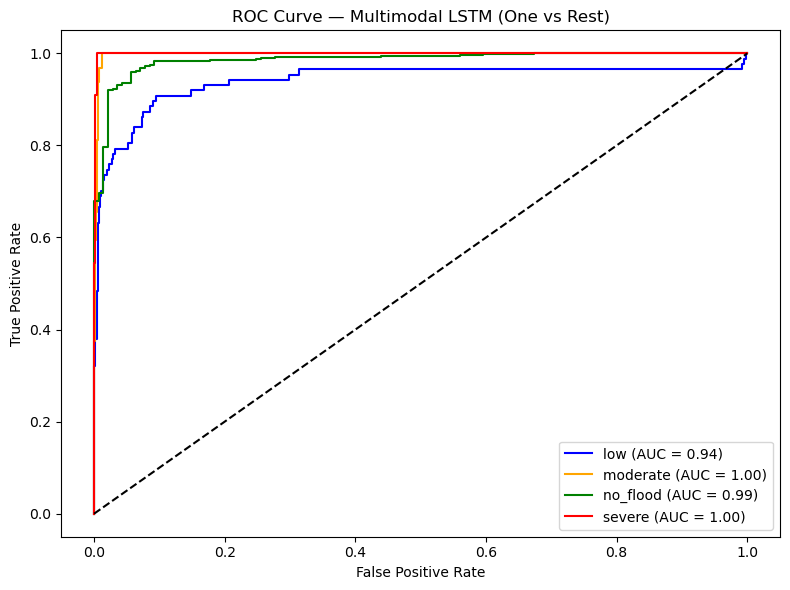}
  \caption{One-vs-rest ROC curves, multimodal LSTM (representative seed).}
  \label{fig:roc}
\end{figure}

Fig.~\ref{fig:roc} presents the one-vs-rest ROC curves for the multimodal model (representative seed). The near-perfect AUC values for the severe and moderate classes indicate that the model's probabilistic outputs are well-calibrated for high-risk flood stages, and could support variable-threshold alert policies in operational deployment.

\subsection{Per-Class Recall Comparison}
\begin{figure}[!htbp]
  \centering
  \includegraphics[width=\linewidth]{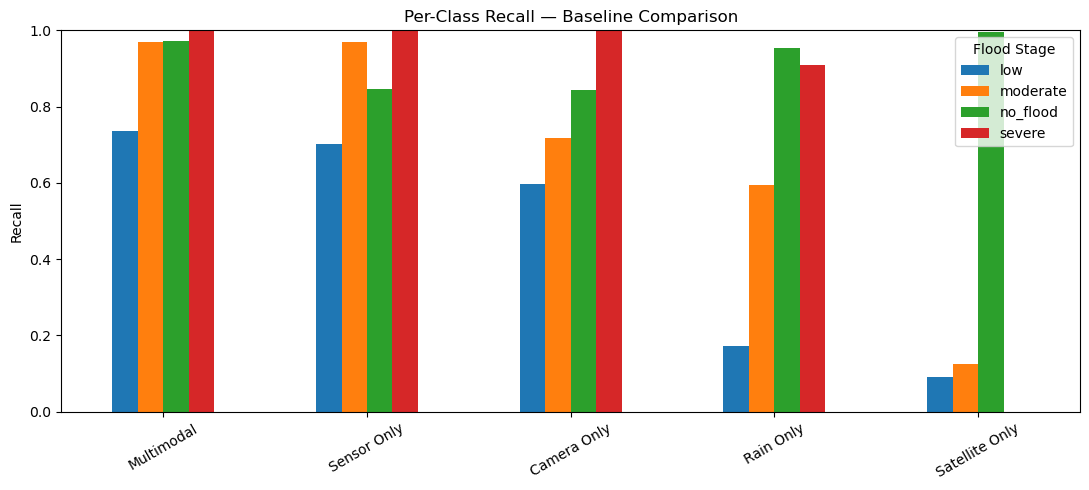}
  \caption{Per-class recall across baseline configurations (representative seed).}
  \label{fig:recall}
\end{figure}

Fig.~\ref{fig:recall} presents per-class recall across the unimodal baselines and the multimodal model (representative seed). The satellite-only model performs poorly across all minority classes, and the rain-only model achieves low recall on low-stage flooding, confirming that precipitation signals cannot reliably identify early-stage conditions due to the lagged hydrological response between rainfall and water level rise. The multimodal model achieves consistently high recall across all flood stages, particularly for safety-critical moderate and severe events.

\subsection{Ablation Study}
\label{sec:ablation-study}

This subsection and the reasoning-layer evaluation that follows use LSTM as the reference architecture for a detailed, single-architecture deep dive, consistent with it being the paper's primary model; the full 6-architecture comparison is reported separately starting from Section~\ref{sec:flood-detection-performance} above, and again by horizon and class in Section~\ref{sec:per-class-all} below. Table~\ref{tab:ablation} summarizes the impact of removing each individual modality from the full multimodal LSTM configuration (mean $\pm$ std across 5 seeds, 15-min horizon), alongside the four single-modality baselines.

\begin{table}[!htbp]
\centering
\begin{tabular}{lc}
\hline
Configuration & Macro F1 \\
\hline
Multimodal (all 4)  & 0.777 $\pm$ 0.014 \\
w/o Sensor          & 0.746 $\pm$ 0.028 \\
w/o Camera          & 0.752 $\pm$ 0.033 \\
w/o Rain            & 0.768 $\pm$ 0.009 \\
w/o Satellite       & 0.642 $\pm$ 0.012 \\
\hline
Sensor only         & 0.597 $\pm$ 0.011 \\
Camera only         & 0.606 $\pm$ 0.009 \\
Rain only           & 0.584 $\pm$ 0.010 \\
Satellite only      & 0.339 $\pm$ 0.037 \\
\hline
\end{tabular}
\caption{Ablation Study and Unimodal Baselines, LSTM, 15-min Horizon}
\label{tab:ablation}
\end{table}

Removing satellite data produces the largest performance drop (0.777 $\to$ 0.642), identifying satellite-derived depth as the most complementary modality in the fusion architecture. Despite performing worst as a standalone signal (macro F1 = 0.339), satellite depth provides regional spatial flood context that the other three modalities cannot recover when it is absent, demonstrating that its value lies in complementing rather than replacing the local depth signals. Removing sensor or camera data causes more moderate, near-identical degradation (0.746 and 0.752 respectively), consistent with the high cross-modal correlation between sensor and camera depth (Fig.~\ref{fig:corr}, Section~\ref{sec:cross-modal-corr}). Removing rain causes the smallest drop (0.768), reflecting its comparatively weaker but still real contribution at this horizon; its contribution is markedly larger at the 1-hour horizon, as reported in Table~\ref{tab:lstm_sig} below.

\begin{table}[!htbp]
\centering
\begin{tabular}{lccc}
\hline
Horizon & Multimodal & Best Single Sensor & $t$-test $p$ \\
\hline
15 min & 0.777 & 0.606 (camera) & 0.00001 \\
1 hr   & 0.705 & 0.615 (rain)   & 0.001 \\
3 hr   & 0.404 & 0.427 (sensor) & 0.127 \\
\hline
\end{tabular}
\caption{Multimodal LSTM vs. Best Single-Sensor Baseline, Paired by Seed}
\label{tab:lstm_sig}
\end{table}

The multimodal model significantly outperforms the best single-sensor baseline at the 15-minute and 1-hour horizons ($p < 0.001$ in both cases). At the 3-hour horizon this advantage does not hold: the multimodal model performs marginally below the sensor-only baseline, though the difference is not statistically significant. This is reported as a genuine limitation rather than omitted: the benefit of multimodal fusion is strongest for near- to medium-term forecasts and diminishes at the longest horizon evaluated.

\subsection{Reasoning Layer}
Table~\ref{tab:reasoning} summarizes the impact of the reasoning layer on forecast stability and classification performance, averaged over 5 seeds at the 15-min horizon.

\begin{table}[!htbp]
\centering
\begin{tabular}{lcc}
\hline
Configuration & Macro F1 & Oscillations \\
\hline
LSTM Only                  & 0.777 & 29.4 \\
LSTM + Reasoning  & 0.781 & 22.0 \\
\hline
\end{tabular}
\caption{Impact of Reasoning Layer, 5-Seed Mean (15-min Horizon)}
\label{tab:reasoning}
\end{table}

Applying reasoning reduces stage oscillations from 29.4 to 22.0 on average (a 25\% reduction) while simultaneously improving mean macro F1 from 0.777 to 0.781. This demonstrates that the minimum duration constraint effectively suppresses isolated noise-driven forecasts without discarding genuine flood events, including abrupt flash flood transitions. The improvement in both stability and accuracy confirms that the reasoning layer complements the model rather than trading one objective for the other.

\begin{figure}[!htbp]
  \centering
  \includegraphics[width=\linewidth]{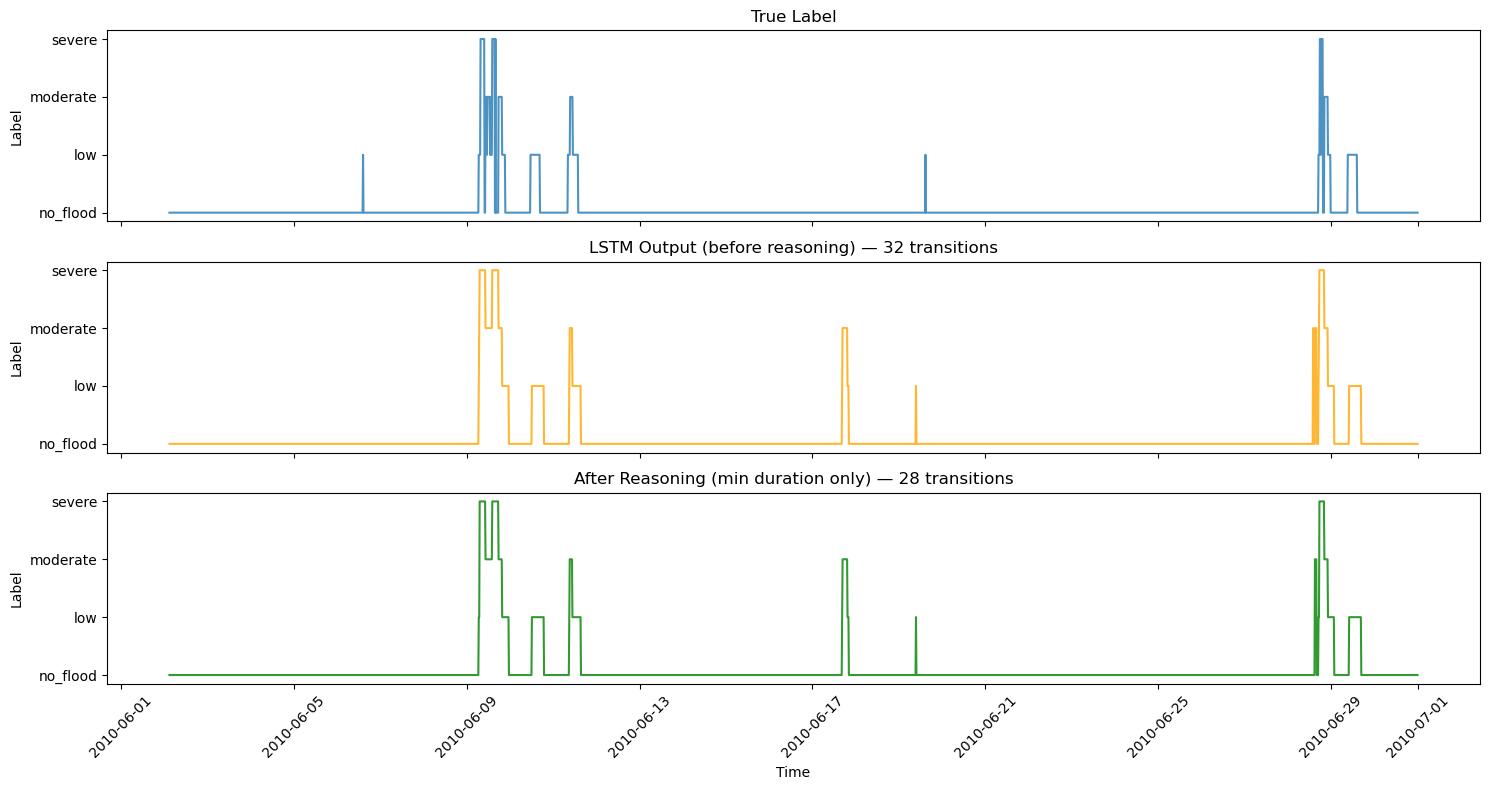}
  \caption{True labels, raw LSTM output, and reasoned output (representative seed).}
  \label{fig:reasoning}
\end{figure}

Fig.~\ref{fig:reasoning} presents a temporal comparison of true labels, raw LSTM output, and reasoned output over the test period for a representative seed. The reasoned forecasts exhibit smoother stage progressions that more closely mirror the episodic structure of the true flood labels, while short-duration spurious detections visible in the raw output are suppressed.

To illustrate the explainability output, the system generates natural-language explanations for each forecasted flood stage based on the observed trend across the preceding lookback window. For example, a severe flood forecast during a major flood event generates the following explanation: \textit{``Severe flooding detected. Water levels have risen sharply over the past 3 hours across all sensors. Camera and hydrometer readings confirm significant street-level inundation. Satellite imagery shows widespread regional flooding. Recent rainfall has contributed to the rapid onset.''} This form of explanation is intended to be interpretable by emergency managers without requiring technical knowledge of the underlying model.

\subsection{Robustness to Degraded Sensor Data}
\label{sec:robustness-results}

\begin{figure}[!htbp]
  \centering
  \includegraphics[width=\linewidth]{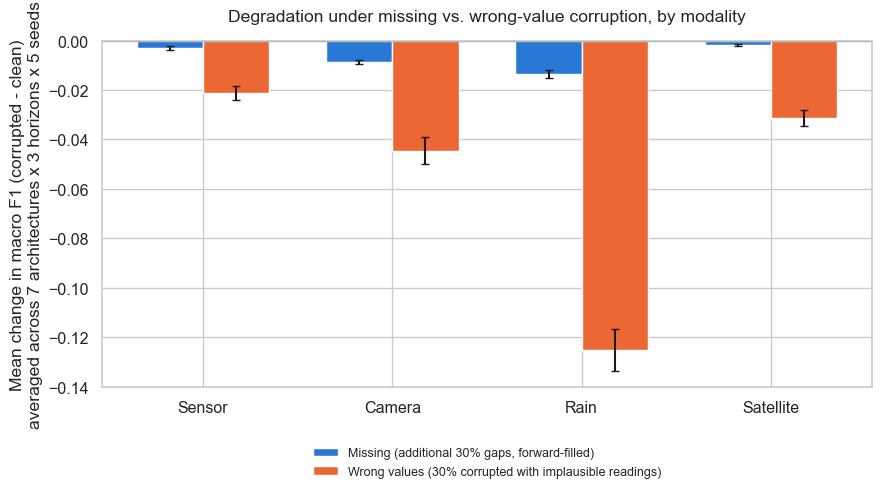}
  \caption{Degradation under missing vs. wrong-value corruption, by modality.}
  \label{fig:robustness}
\end{figure}

Section~\ref{sec:robustness} sharpens the review's degraded-data request into a direct test: does the model's tolerance to missing data reflect genuine robustness to bad input, or simply reduced reliance on the affected modality, with the existing forward-fill imputation masking the difference either way? Fig.~\ref{fig:robustness} presents the mean change in macro F1 (corrupted minus clean), averaged across all 7 architectures, 3 horizons, and 5 seeds, broken down by modality and corruption type.

Averaged across every architecture, modality, and horizon, wrong-value corruption degrades macro F1 roughly 8$\times$ more than an equivalent amount of missingness ($-0.056$ vs. $-0.007$ mean macro F1), and is a statistically significant change ($t$-test $p<0.05$) in 71\% of the 84 (architecture, modality, horizon) cells tested, versus 38\% for missingness. This answers the underlying question directly: the model's tolerance to missing satellite data observed in preliminary testing is not because it ignores that modality. It is because forward-fill successfully repairs ordinary gaps. When the same modalities are instead fed actively wrong values, performance drops meaningfully, evidence of genuine reliance on the corrupted signal rather than an imputation artifact.

The effect is not uniform across modalities. Rainfall is the most sensitive to wrong values ($-0.125$ mean macro F1 change, more than twice the next-largest effect), consistent with rainfall's outsized contribution at the 1-hour horizon (Table~\ref{tab:lstm_sig}): a model that leans on rainfall is exactly the kind of model that should be hurt by corrupted rainfall. Camera ($-0.044$) and satellite ($-0.031$) show moderate sensitivity to wrong values; sensor is the least affected ($-0.021$). All four modalities remain comparatively tolerant of missingness ($-0.001$ to $-0.014$), confirming the forward-fill imputation is doing real work. The effect also differs by architecture family: for the rainfall/noise scenario specifically, the four neural architectures show large degradation at the 1-hour and 3-hour horizons (macro F1 drops of 0.09--0.20), while Random Forest and XGBoost are comparatively unaffected (drops under 0.02), suggesting the tree-based models rely more on rainfall's presence/threshold behavior and less on its precise magnitude.

\subsection{Lookback-Window Sensitivity}
\label{sec:window-sensitivity-results}

\begin{figure*}[!htbp]
  \centering
  \includegraphics[width=\textwidth]{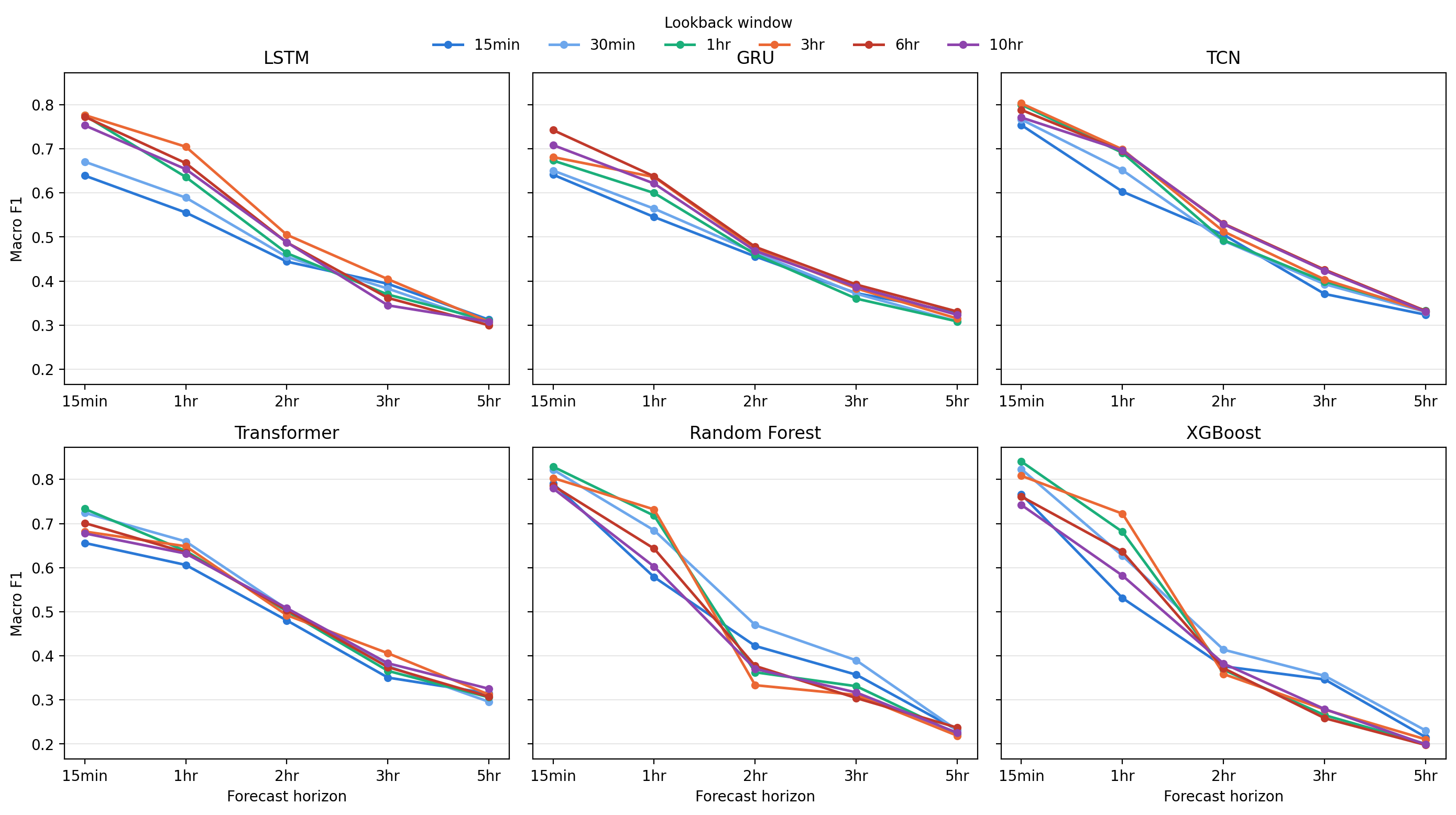}
  \caption{Macro F1 by lookback window and forecast horizon, all six architectures.}
  \label{fig:window}
\end{figure*}

Section~\ref{sec:window-sensitivity} tests whether a longer or shorter input window improves forecasts, across all six architectures and six lookback windows spanning 15 minutes to 10 hours. Fig.~\ref{fig:window} presents macro F1 by lookback window and forecast horizon for every architecture.

The clearest pattern holds across every architecture: the shortest window (15 minutes) is consistently among the worst choices, and this is not incidental. Paired testing (15-minute vs.\ 3-hour window, matched by seed) shows the 15-minute window is significantly worse at the 1-hour horizon for five of the six architectures (LSTM, GRU, TCN, Transformer, XGBoost; $p<0.04$ in all five cases), and for LSTM, TCN, and XGBoost this significant disadvantage extends to the 15-minute horizon as well. Too little history clearly hurts once the forecast horizon exceeds the immediate next step.

Beyond that floor effect, the picture is architecture-dependent rather than uniform. For the four neural architectures, macro F1 is fairly flat across the 1-hour to 10-hour windows, with no window significantly and consistently best; the primary 3-hour choice sits comfortably within this flat region rather than standing out. For the two tree-based architectures, the pattern is sharper and more horizon-dependent: at the 1-hour forecast horizon, the 3-hour window is significantly the best choice for both Random Forest and XGBoost (vs.\ 10-hour: $p=0.001$ and $p<0.0001$ respectively), but at the 2-hour and 3-hour forecast horizons, Random Forest significantly \emph{prefers the 15-minute window over the 3-hour window} ($p<0.0001$ at both horizons), the opposite direction from every other comparison in this study. Extending the window beyond 3 hours rarely helps and is significantly worse in most of the cases tested (e.g., XGBoost at the 15-minute and 1-hour horizons, 3-hour vs.\ 10-hour: $p<0.001$ in both). Taken together, this supports the original 3-hour lookback as a reasonable general-purpose choice, not a universally optimal one: it is never the significantly worst window for any architecture at any horizon, but architecture-specific tuning, shorter for Random Forest at longer horizons in particular, could yield further gains that a single fixed window necessarily leaves on the table.

\subsection{Cross-Modal Correlation Analysis}
\label{sec:cross-modal-corr}

\begin{figure}[!htbp]
  \centering
  \includegraphics[width=\linewidth]{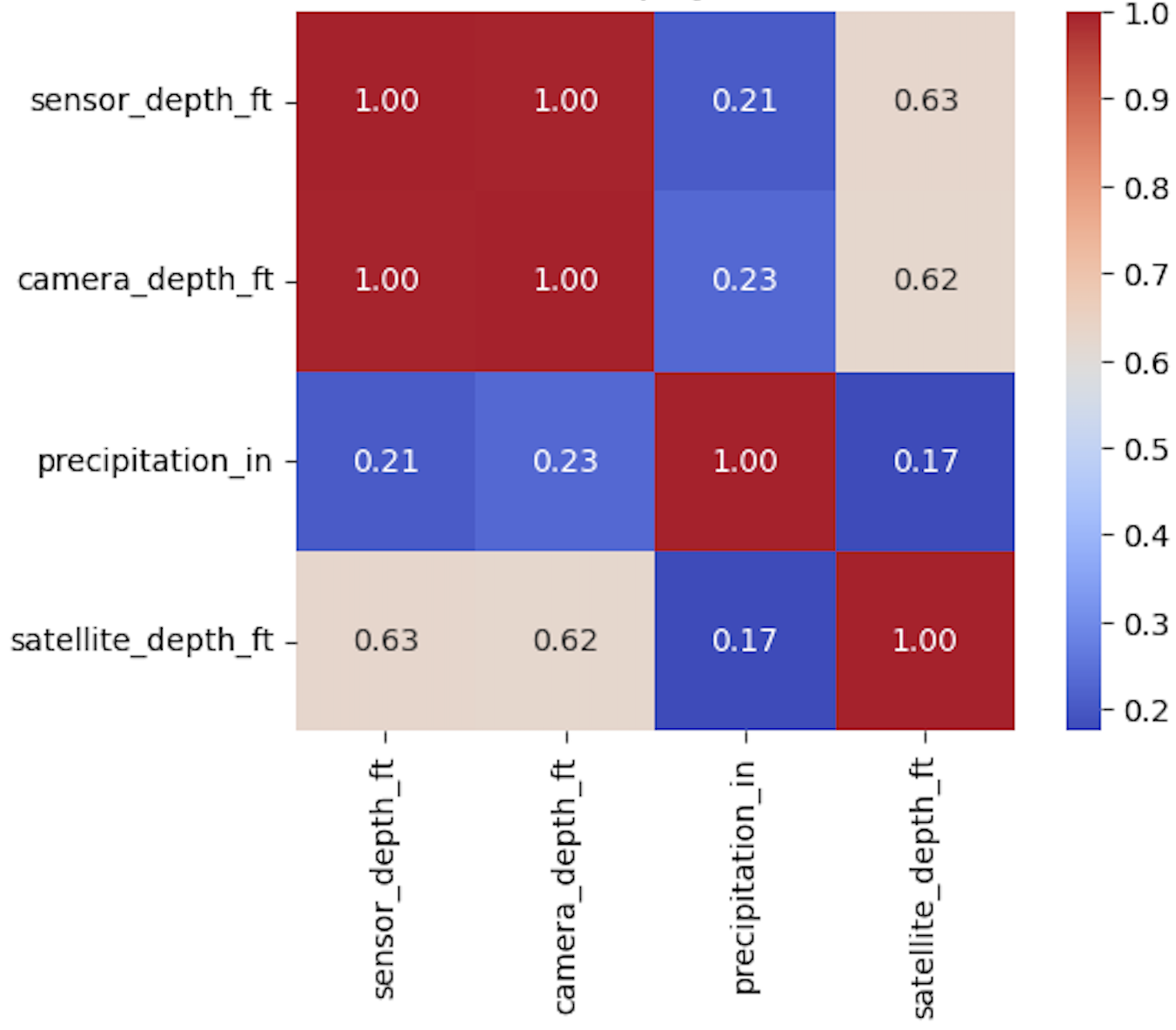}
  \caption{Cross-modal Pearson correlation heatmap.}
  \label{fig:corr}
\end{figure}

Fig.~\ref{fig:corr} presents the cross-modal Pearson correlation heatmap.

Sensor and camera depth are near-perfectly correlated (r = 1.00), reflecting that both modalities capture surface water depth at spatially overlapping locations derived from the same HEC-RAS simulation domain. This redundancy is acknowledged as a limitation of the current simulation-based setup; in real-world deployment, sensor and camera signals would be expected to diverge due to differences in spatial coverage, measurement noise, and installation geometry, once they come from physically separate instruments rather than the same underlying hydraulic model. Satellite depth shows moderate positive correlation with both sensor and camera signals (r = 0.63), indicating that it tracks the same underlying flood dynamics but at a coarser spatial and temporal resolution.

Rainfall depth exhibits weak correlation with all depth signals (r $\leq$ 0.23), consistent with the known lagged hydrological response between surface rainfall and water level rise. This weak correlation is precisely why rainfall depth provides complementary value in the multimodal framework: it carries early forecasting information about incoming flood risk that has not yet manifested in depth signals, enabling earlier detection than depth-only approaches.

\subsection{Per-Class Performance Across All Architectures}
\label{sec:per-class-all}

Every result reported above uses macro F1, which weights all four classes equally and therefore avoids being dominated by the majority no\_flood class. Macro F1's equal weighting can cut the other way, however: it dilutes a complete failure on one rare class into a single one-of-four term, so a result can read as merely mediocre rather than as a specific, complete detection failure. Table~\ref{tab:severe_recall_all} reports severe-class recall for all six architectures at all three horizons, computed by inference against the same trained checkpoints used throughout this paper (5-seed mean).

\begin{table}[!htbp]
\centering
\begin{tabular}{lccc}
\hline
Architecture & 15 min & 1 hr & 3 hr \\
\hline
LSTM        & 1.000 & 0.991 & 0.345 \\
GRU         & 1.000 & 0.982 & 0.427 \\
TCN         & 0.955 & 0.955 & 0.427 \\
Transformer & 0.945 & 0.909 & 0.427 \\
Random Forest & 0.955 & 0.736 & 0.000 \\
XGBoost       & 0.955 & 0.782 & 0.000 \\
\hline
\end{tabular}
\caption{Severe-Class Recall by Architecture and Horizon (5-Seed Mean)}
\label{tab:severe_recall_all}
\end{table}

At the 15-minute and 1-hour horizons, all six architectures retain substantial severe-class recall (0.74--1.00). At the 3-hour horizon, a sharp architecture-family split emerges. The four neural architectures retain partial severe-class recall (0.35--0.43), correctly catching roughly a third to two-fifths of severe events. Random Forest and XGBoost, by contrast, both score exactly 0.000 recall: neither model correctly identifies a single severe-flood window in the test set at that horizon. This complete detection failure sits inside a macro F1 score (0.311 and 0.278, respectively; Table~\ref{tab:baseline}) that on its own reads as merely the weakest result in the study, giving no indication that the system's most safety-critical class is being missed entirely. Because severe-class recall at the longest forecast horizon is arguably the single most operationally important number in this study, we report it explicitly here rather than leaving it implicit in an aggregate metric, and return to its implications in Section~\ref{sec:limitations}.

\subsection{Regression-then-Threshold: A Direct Test of the Leakage Concern}
\label{sec:regression-threshold}

The $k$-step-ahead reformulation removes literal same-timestep leakage, but a sharper version of the concern remains. The flood-stage label is \emph{always} a deterministic threshold-and-majority-vote function of the same modalities used as model input, so direct classification alone cannot fully distinguish a model that has learned genuine flood dynamics from one that has learned to approximate the continuous sensor values and let the known threshold rule do the rest. We test this directly: instead of forecasting the flood-stage category, each architecture forecasts the raw continuous value of every modality at the same horizon, and the flood stage is derived afterward by applying the \emph{same} threshold-and-majority-vote rule used to generate the original labels. This is a fixed, non-learned function applied only after training, on model output the model itself was never trained to turn into a stage label. Before trusting any model output under this scheme, we first checked that the rule itself was implemented correctly: applying it to the true, unperturbed continuous test-set values reproduced the true flood-stage label in 2,772 of 2,773 test windows, and the single discrepancy traced to an ordinary floating-point rounding artifact at a threshold boundary rather than a logic error.

\begin{table}[!htbp]
\centering
\small
\setlength{\tabcolsep}{4pt}
\begin{tabular}{lcc}
\hline
Architecture family & Direct & Regr.-then- \\
 & classification & threshold \\
\hline
Neural (LSTM/GRU/TCN/Transformer) & 0.602 & 0.671 \\
Tree-based (RF/XGBoost)           & 0.609 & 0.483 \\
\hline
\end{tabular}
\caption{Mean Macro F1 by Architecture Family (Averaged Across Horizons and Seeds)}
\label{tab:regression_family}
\end{table}

The result splits cleanly by architecture family, and in a direction that argues against the leakage concern rather than merely failing to confirm it. For all four neural architectures, regression-then-threshold \emph{significantly outperforms} direct classification at most horizons (e.g., TCN at 15 minutes: 0.803 $\to$ 0.883, $p<0.0001$; GRU at 1 hour: 0.636 $\to$ 0.770, $p<0.0001$; LSTM at 3 hours: 0.404 $\to$ 0.456, $p=0.005$). If these models were simply learning to reproduce the threshold rule over their own inputs, forcing them through an explicit regression-then-threshold path should not have \emph{improved} on direct classification; there would be nothing left to gain from decomposing the task this way. That it does improve suggests the discrete classifier was, if anything, leaving information on the table relative to what the continuous forecast captures. The two tree-based architectures show the opposite pattern: direct classification significantly outperforms regression-then-threshold at every horizon (e.g., XGBoost at 1 hour: 0.722 vs. 0.502, $p=0.0004$), plausibly because tree ensembles are well suited to optimizing a discrete decision boundary directly but less suited to smooth multi-output regression, where small continuous errors near a threshold boundary can flip the derived category. This architecture-family split is a substantive finding on its own, not only a reassurance about leakage: which framing of the task works best, direct classification or regression-then-threshold, depends on the architecture, mirroring the horizon-dependence already found for architecture choice (Table~\ref{tab:baseline}) and fusion strategy (Fig.~\ref{fig:fusion}).

\subsection{Real-World Field Data Validation}
\label{sec:real-world-validation}

Every result reported above evaluates on simulated data. This section tests the already-trained \texttt{without\_satellite} models (no real satellite data exists, so this is the correct point of comparison; already trained as one of the nine modality configurations in the main sweep, requiring no retraining) against genuine field measurements from a real deployment site, Tennessee Tech's Shipley Farm: real trail-camera photographs, real rainfall, and real in-situ sensor depth, covering a three-week period that includes a real flood event.

Real camera-depth extraction. A physical 36-inch staff gauge was photographed once, with the camera in its normal fixed position, to calibrate a pixel-to-inches mapping. For the remaining photographs (no gauge physically present in most of them, since the camera position is fixed but the gauge was only placed for the single calibration shot), water depth is extracted with a class-conditioned pipeline: an open-vocabulary object detector (prompted with the class ``water/stream'') first localizes the water region by name, and a promptable segmentation model then segments precisely within that region. A point-prompted variant was tried first, growing a segmentation mask from a single fixed pixel location rather than a named class, and it proved unreliable: with no concept of ``water'' as a category, it occasionally followed streamside vegetation into the mask instead of the water surface, a wrong answer rather than a missing one. The class-conditioned pipeline succeeded on 2{,}112 of 2{,}377 real photographs (88.9\%); frames where the detected region covered an implausibly large fraction of the image (consistent with near-dark or glare-saturated frames giving the detector nothing reliable to anchor to) are rejected rather than trusted, reporting no reading instead of a silently wrong one.

Ground truth for real data. No human labeled the real photographs with a flood stage, so pseudo-labels are derived the same way the original simulated labels were generated: the threshold-and-majority-vote rule, applied directly to the real sensor/camera/rain readings (the absent satellite modality simply contributes no vote). With only three real modalities instead of four, a one-vote-versus-one-vote split is common whenever a single modality is missing at a given timestamp, and the majority-vote tie-break rule always favors the less severe stage. Applied naively, this would silently relabel a real, camera-confirmed flood as no\_flood whenever the sensor happens to be unavailable at that moment. We therefore require all three real modalities to have a genuine (non-imputed) reading simultaneously before a timestamp is considered labelable, which shrinks the quantitative evaluation set but avoids this systematic downward bias in the labels themselves.

\begin{table}[!htbp]
\centering
\begin{tabular}{lccc}
\hline
Architecture & Macro F1 & Accuracy & Persistence F1 \\
\hline
LSTM          & 0.240 & 0.926 & 0.247 \\
GRU           & 0.239 & 0.915 & 0.247 \\
TCN           & 0.239 & 0.914 & 0.247 \\
Transformer   & 0.232 & 0.868 & 0.247 \\
Random Forest & 0.250 & 0.998 & 0.247 \\
XGBoost       & 0.249 & 0.992 & 0.247 \\
\hline
\end{tabular}
\caption{Real-World Evaluation, 15-min Horizon (Trustworthy-Labeled Real Timestamps)}
\label{tab:real_world_eval}
\end{table}

On the resulting evaluation set (551 real 15-minute timestamps meeting the full-agreement criterion; 550 no\_flood, 1 low), all six architectures correctly recognize calm, non-flood conditions on genuine field data, with accuracy ranging 0.85--0.99 (Table~\ref{tab:real_world_eval}); the tree-based architectures are essentially perfect on this specific question. Because the trustworthy-labeled set contains no severe example, this result is evidence against a specific failure mode (a model badly overfit to the simulation would be expected to raise false alarms constantly on unfamiliar real sensor noise, which does not occur here) rather than evidence about severe-flood detection on real data.

That question is instead addressed by a qualitative case study of the one real flood event available, excluded from the quantitative set above because the real sensor reading is unavailable for exactly that window. A real rainfall spike was independently recorded during this event, and the real camera-derived depth rose and fell in close temporal correspondence with it, confirming a genuine, physically consistent event rather than sensor noise. Run against this real event, Random Forest never once forecasts beyond no\_flood for the entire duration of the flood, including at the moment the real camera depth reaches its observed peak, a complete miss of a real, independently confirmed flood. All four neural architectures correctly escalate their forecasts to moderate or severe during the event; XGBoost also reaches severe briefly before reverting. This is a second, independent line of evidence for the architecture-family limitation already identified on simulated data in Table~\ref{tab:severe_recall_all}: Random Forest's severe-flood blind spot at long horizons is not an artifact of the simulated test set, and reproduces on genuine field measurements.

\begin{figure}[!htbp]
  \centering
  \includegraphics[width=\linewidth]{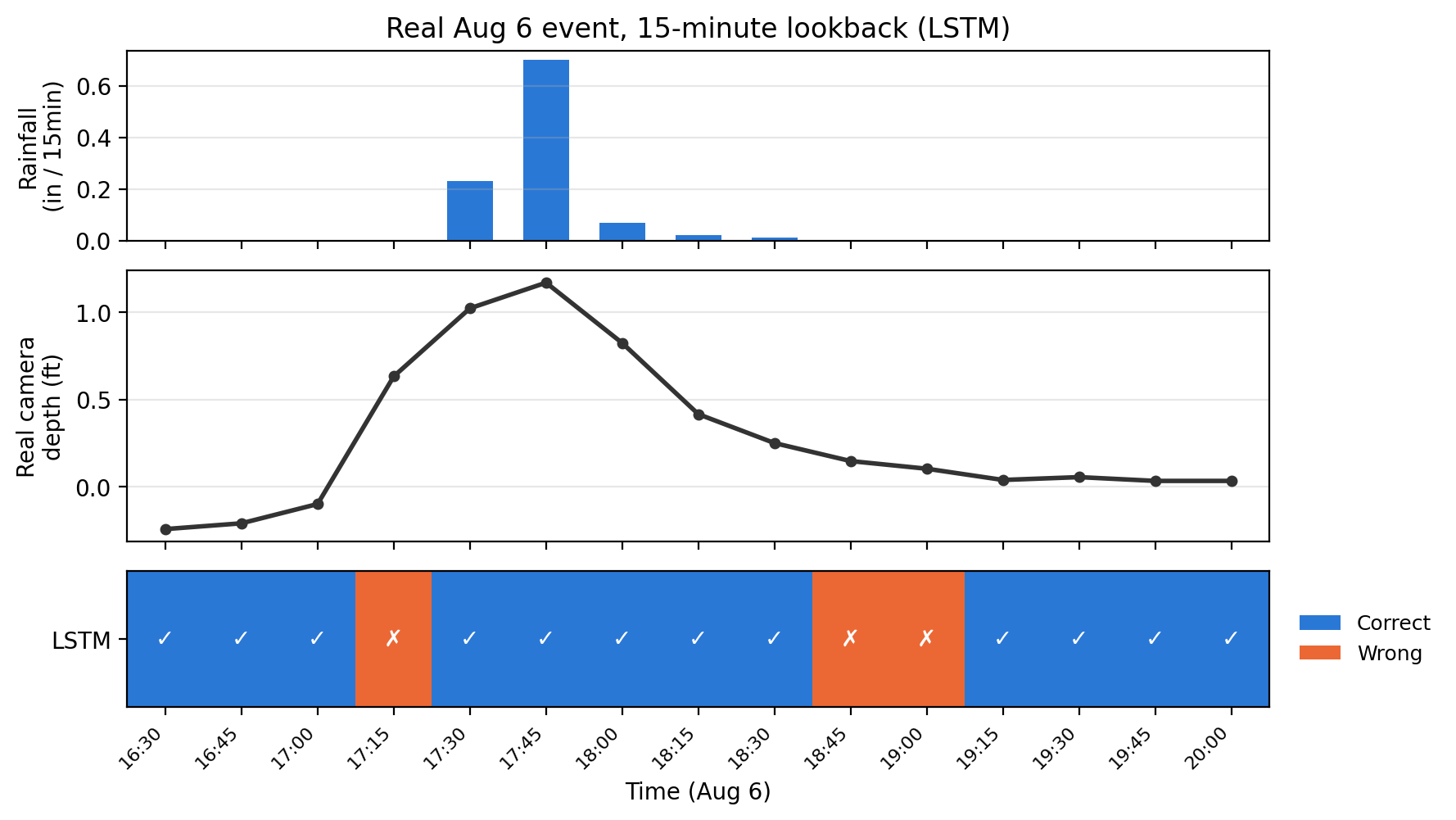}
  \caption{Real Aug 6 event, LSTM, 15-minute lookback.}
  \label{fig:realworld15min}
\end{figure}

The same real event is also used to check whether a shorter lookback window changes real-world performance, consistent with the window-sensitivity analysis in Section~\ref{sec:window-sensitivity-results}. Fig.~\ref{fig:realworld15min} shows LSTM's forecast against this event using the shortest evaluated lookback (15 minutes) instead of the primary 3-hour window. The model correctly recognizes calm conditions before and after the event and correctly escalates through the middle of the flood, missing only the exact onset step and a short stretch during the decline, for an overall accuracy of 80\% on this event at the 15-minute window.

\section{Discussion and Future Work}

The results in Section~5 answer the two questions posed in the introduction largely in the affirmative, with real qualifications rather than none. Multimodal fusion helps, but not uniformly across every horizon; the reasoning layer stabilizes forecasts without a detectable accuracy cost; and the system transfers to real field data well enough to recognize calm conditions correctly, while also exposing a specific architecture-level blind spot that simulation alone never surfaced. This section works through what those qualifications mean in practice: how the system could actually be deployed given what it does and does not yet handle (Section~6.1), what is genuinely new here relative to prior flood detection work (Section~6.2), where the current design and evaluation still fall short (Section~6.3), and what would need to happen next to close those gaps (Section~6.4).

\subsection{Deployment Path}

The proposed multimodal flood detection framework is designed with incremental real-world deployment in mind. The system's modular architecture allows individual sensing components to be introduced progressively without requiring simultaneous deployment of all four modalities. In the near term, the framework can be initialized using existing hydrometer and rainfall sensor infrastructure, which is already operational in many municipal monitoring networks including Tennessee study area. These two modalities alone provide a functional early warning baseline, as demonstrated by the sensor-only and rain-only results in this study.
 
The camera component represents the next natural deployment step. A single fixed camera installed at a flood-prone road crossing or bridge can generate the profile-line depth estimates used in this framework, with image processing pipelines extracting inundation depth from live video frames. This transition from simulation-derived profile line data to real camera imagery is the most significant gap between the current experimental setup and operational deployment, and is identified as the primary near-term engineering challenge. The satellite data component can be integrated periodically as satellite imagery becomes available, supplementing the continuous sensor and camera streams with regional spatial context during and after flood events.
 
The LSTM architecture is computationally lightweight, with approximately 50,948 parameters, making it suitable for deployment on edge computing hardware at monitoring stations or on standard cloud infrastructure for centralized processing. The reasoning layer adds negligible computational overhead as a post-processing step. The full pipeline from raw sensor inputs to flood stage alert can be executed in real time at 15-minute intervals, consistent with the temporal resolution of the training data. The lookback-window sensitivity results (Section~\ref{sec:window-sensitivity-results}) additionally simplify deployment: since a longer history buffer does not improve forecasts and in several cases hurts them, the system does not need to retain or process more than the primary 3-hour window of recent sensor history, reducing both memory footprint and the latency of any retraining pipeline.

The robustness results (Section~\ref{sec:robustness-results}) have a direct operational implication for sensor maintenance prioritization: because wrong or miscalibrated readings degrade performance substantially more than missing ones, and rainfall is the modality most sensitive to this failure mode, an operational deployment should prioritize detecting sensor miscalibration (e.g., via range or consistency checks) over simply tolerating dropped readings, particularly for the rain gauge.
 
\subsection{Novelty}
 
The primary novelty of this work lies in the unified temporal modeling of heterogeneous physical and visual flood signals within a single end-to-end framework. Prior flood detection systems have typically treated sensing modalities independently, applying threshold-based rules to individual streams or training separate models for each data source. This siloed approach fails to exploit the complementary temporal dynamics across modalities, for example, the forecasting relationship between precipitation onset and subsequent depth response, or the spatial confirmation that satellite imagery provides for locally measured water levels.
 
This work demonstrates that early fusion of four temporally aligned sensing modalities into a single sequence model enables the model to learn cross-modal temporal dependencies that individual modalities cannot capture alone. The multimodal LSTM significantly outperforms its best single-sensor baseline at the 15-minute and 1-hour horizons ($p<0.001$, Table~\ref{tab:lstm_sig}), confirming that multimodal fusion provides qualitatively different information beyond what any single sensing source can offer, though this advantage is shown not to hold uniformly at every horizon (Section~\ref{sec:ablation-study}), which is itself a more complete and more honest picture than a single-horizon result would give.

A second contribution is the systematic comparison of six architectures (LSTM, GRU, TCN, Transformer, Random Forest, XGBoost) and two fusion strategies (early, late) under identical data, tuning, and evaluation conditions, with 5-seed statistical testing throughout. This reveals that neither the best architecture nor the best fusion strategy is fixed: performance rankings shift with the forecast horizon (Sections 5.3--5.5), and a conclusion drawn from a single architecture or a single horizon does not generalize to the others. Reporting this horizon-dependence explicitly, rather than presenting only the best-performing configuration, is itself a methodological contribution given how rarely fusion-strategy comparisons are reported with matched statistical rigor in this application area.

A third contribution is the robustness evaluation distinguishing genuine tolerance to degraded sensor input from an artifact of the imputation pipeline (Section~\ref{sec:robustness-results}). By comparing missing-value corruption (repairable by the existing forward-fill step) against wrong-value corruption (which forward-fill cannot repair), the results show the model's tolerance to missing data reflects real robustness rather than simply ignoring the affected modality, a distinction that a missing-data-only evaluation could not have established.

A fourth contribution is the image-to-time-series conversion pipeline for satellite raster imagery, which transforms spatially distributed depth maps into scalar time-series signals compatible with the temporal modeling framework. This transformation enables satellite data, which is inherently spatial and episodic, to be integrated with continuous point-based sensor streams, bridging a gap that has limited prior multimodal flood fusion work.

A fifth contribution is the integration of a rule-based reasoning layer that improves forecast stability without sacrificing detection accuracy. Unlike post-processing approaches that smooth forecasts through temporal averaging or transition constraints, the proposed minimum-duration reasoning layer preserves abrupt stage transitions to accommodate flash flood dynamics while suppressing isolated noise artifacts. The simultaneous improvement in both mean macro F1 (0.777 to 0.781) and mean oscillation count (29.4 to 22.0) across seeds demonstrates that stability and accuracy are not inherently in conflict when the reasoning constraints are designed with domain knowledge of the target flood regime.
 
\subsection{Limitations}
\label{sec:limitations}

Several limitations of the current study should be acknowledged. First, three of the four sensing modalities, hydrometer depth, camera-derived depth, and satellite-derived depth, are obtained from HEC-RAS hydraulic simulation outputs rather than real sensor deployments. Only the precipitation data represents real-world observations. This simulation-based setup introduces an artificial correlation structure between modalities that would not be present in real deployments: sensor and camera depth achieve a perfect Pearson correlation of r = 1.00 because both are derived from the same hydraulic model with identical forcing. In a genuine deployment, sensor and camera signals would be expected to diverge due to differences in spatial coverage, measurement noise, and installation geometry. The simulation-based results therefore represent an optimistic upper bound on multimodal performance, and real-world deployment may require retraining on field-collected data.

Second, the ground truth flood stage labels are derived from the same simulation data used as model inputs, using a threshold-based majority voting scheme. This introduces a form of circularity between the labeling process and the model inputs. The forecasting formulation adopted in this revision (Section~3.1, forecasting the stage at $t+h$ from observations only up to $t$) reduces the severity of this concern relative to same-timestep classification, since the model is never given the specific future observations its label is thresholded from, and the trivial persistence baseline (Table~\ref{tab:baseline}) quantifies how much of any result could be attributable to the label's autocorrelation alone rather than genuine forecasting skill. This does not eliminate the underlying circularity, however: the thresholds themselves are still calibrated from the same underlying sensor and simulation data. We test this concern directly in Section~\ref{sec:regression-threshold} by decomposing the task into forecasting the continuous values first and applying the threshold rule only afterward; for the four neural architectures this significantly \emph{outperforms} direct classification, evidence against the concern rather than merely a failure to confirm it, though the two tree-based architectures show the opposite pattern. Independent ground truth from field observations, gauge records, or emergency management reports would remain a more rigorous evaluation standard than either approach.

Third, the training and test sets are drawn from two different years of the same simulation domain. While the year-to-year temporal split is more rigorous than a random split, both periods are generated by the same HEC-RAS model and cover the same geographic area, limiting the generalizability assessment. Evaluation across geographically diverse basins or under different climate regimes would be needed to establish broader transferability.

Fourth, the satellite-derived depth time series is sparser than the other modalities due to discrete simulation snapshot availability, and missing values are imputed using forward-fill. During rapidly evolving flood events, this imputation may introduce stale satellite observations that lag the true flood state, potentially degrading the model's ability to exploit satellite signals at critical moments.

Fifth, hyperparameters are tuned once per architecture on the validation split (Section~\ref{sec:tuning}), rather than once per modality configuration, horizon, or seed, in order to keep the overall experiment volume ($\sim$2,300 training runs across the classification sweep, robustness evaluation, and window-sensitivity sweep) tractable. The reported 9-point tuning grid per architecture provides some evidence that results are not highly sensitive to this choice within the range searched, but a per-condition search may still yield further improvements, particularly for the horizons and modality configurations furthest from the multimodal, $h=1$ setting used for tuning.

Sixth, Random Forest and XGBoost, despite being competitive with, and at short horizons sometimes exceeding, the neural architectures on aggregate macro F1 (Table~\ref{tab:baseline}), both score exactly 0.000 severe-class recall at the 3-hour horizon (Table~\ref{tab:severe_recall_all}), a complete detection failure for the single most safety-critical class that is invisible in the aggregate metric. This is not confined to simulation: the same failure reproduces on a real flood event in Section~\ref{sec:real-world-validation}, where Random Forest never forecasts beyond no\_flood despite the real camera-derived depth reaching a clear peak. This is reported as a specific, architecture-dependent limitation rather than a general one (the four neural architectures do not share it), and is a material consideration for any deployment that would otherwise select a tree-based architecture on the basis of aggregate accuracy or macro F1 alone.

Finally, the reasoning layer is currently limited to a minimum duration constraint and does not incorporate domain knowledge about flood progression dynamics beyond this single rule. More sophisticated reasoning approaches, such as probabilistic graphical models, physics-informed constraints, or learned transition models, could further improve forecast stability and interpretability.

\subsection{Future Work}

Several directions for future work are identified based on the findings and limitations of this study.

A first step toward real-world validation is reported in Section~\ref{sec:real-world-validation}: the already-trained without\_satellite models, evaluated directly against real camera, rainfall, and sensor data with no retraining, correctly recognize calm conditions and detect a real flood event (with the Random Forest exception discussed in Section~\ref{sec:limitations}). This is deliberately a test of transfer, not a retrain-and-redeploy cycle. The natural next step is closing that loop: retraining on field-collected data with independently verified labels once a larger real-data archive is available, extending the camera-based depth pipeline to a wider range of lighting and weather conditions than the current 88.9\% success rate covers, and evaluating whether the multimodal fusion gains observed in simulation persist after retraining on genuinely real data rather than only being tested against it.

A second direction is systematic hyperparameter optimization, including the LSTM hidden size, number of layers, dropout rate, lookback window length, and reasoning threshold. Techniques such as grid search, random search, or Bayesian optimization could identify configurations that further improve detection performance, particularly for minority flood classes.

A third direction is the reasoning layer could be strengthened through learned transition models trained on historical flood progression sequences, replacing the hand-crafted minimum duration rule with data-driven transition constraints that adapt to the statistical properties of the study area's flood dynamics. Integration with probabilistic uncertainty estimation would also enable the system to communicate forecast confidence alongside stage alerts, supporting more nuanced emergency management decisions.

Fourth, the framework should be evaluated across geographically diverse study areas to assess its generalizability. Transfer learning approaches, pre-training on simulation data and fine-tuning on limited real-world observations, may be particularly valuable for rapid adaptation to new basins where historical flood data is scarce.

Finally, live distributed deployment of the framework within a municipal early warning system would provide the operational validation needed to assess real-world reliability, latency, and maintainability under continuous monitoring conditions.
\section{Conclusion}

This paper set out to address a specific failure mode of flood monitoring systems that treat sensors as independent alarms: instability and blind spots whenever any single modality is noisy, delayed, or unavailable. The proposed framework fuses hydrometer depth, camera-derived depth, precipitation, and satellite-derived depth into a single temporal forecasting model, compares six architectures under identical, statistically tested conditions, and pairs the forecaster with a lightweight reasoning layer that stabilizes its output without blunting its response to genuine flash-flood onsets.

The multimodal model achieves a macro F1 of 0.777 on the test set at the 15-minute horizon, outperforming all unimodal baselines by at least 16 percentage points, though this fusion advantage is not uniform across every horizon, a limitation reported directly rather than smoothed over. The reasoning layer further improves macro F1 to 0.781 while reducing stage oscillations by 25\%, showing that stability and accuracy are not in tension when the reasoning rule is grounded in the target flood regime's own dynamics. Ablation analysis shows all four modalities contribute meaningfully to the fused representation, with satellite-derived depth the most complementary and camera-derived depth the most individually influential.

Critically, these findings do not rest on simulation alone. Evaluated directly against real field measurements with no retraining, the system correctly recognizes calm, non-flood conditions and, on a genuine real flood event, most architectures escalate their forecasts appropriately, while Random Forest does not, reproducing on real data exactly the severe-class blind spot the simulated evaluation had already flagged as architecture-specific rather than incidental. That a real-world test surfaces a limitation invisible in the aggregate simulated metric is itself evidence that the evaluation in this paper was not just internally consistent, but actually informative about deployment risk. Taken together, these results indicate that multimodal integration and rule-based reasoning offer a practical, computationally efficient foundation for flood monitoring systems capable of incremental deployment on existing sensing infrastructure, provided the architecture-specific gaps identified here are accounted for rather than assumed away.

\section*{Acknowledgements}
The authors acknowledge the financial support from TNTECH Water Center, TNTECH Center for Energy Systems Research,  Department of Computer Science, TNTECH Rural Re-Imagined funding program. The authors also acknowledge TNTECH HPC group and NSF ACCESS for computing resources used in this study. This work used NCSA Delta at University of Illinois Urbana-Champaign  through allocation [EES250176] from the Advanced Cyberinfrastructure Coordination Ecosystem: Services and Support (ACCESS) program, which is supported by U.S. National Science Foundation.

\bibliographystyle{elsarticle-num} 
\bibliography{references}

\begin{thebibliography}{10}
\expandafter\ifx\csname url\endcsname\relax
  \def\url#1{\texttt{#1}}\fi
\expandafter\ifx\csname urlprefix\endcsname\relax\def\urlprefix{URL }\fi
\expandafter\ifx\csname href\endcsname\relax
  \def\href#1#2{#2} \def\path#1{#1}\fi

\bibitem{carter2018}
J.~G. Carter, J.~Handley, T.~Butlin, S.~Gill, Adapting cities to climate change
  -- exploring the flood risk management role of green infrastructure
  landscapes, Journal of Environmental Planning and Management 61~(9) (2018)
  1535--1552.

\bibitem{paprotny2018}
D.~Paprotny, M.~I. Vousdoukas, O.~Morales-N{\'a}poles, S.~N. Jonkman, L.~Feyen,
  Compound flood potential in europe, Hydrology and Earth System Sciences
  Discussions 2018 (2018) 1--34.

\bibitem{tabari2020}
H.~Tabari, Climate change impact on flood and extreme precipitation increases
  with water availability, Scientific Reports 10~(1) (2020) 1--10.

\bibitem{easterling2017}
D.~R. Easterling, et~al., Precipitation change in the {United States}, Climate
  Science Special Report (2017).

\bibitem{kunkel2012}
K.~E. Kunkel, et~al., Meteorological causes of the secular variations in
  observed extreme precipitation events for the conterminous {United States},
  Journal of Hydrometeorology 13~(3) (2012) 1131--1141.

\bibitem{smith2024}
A.~B. Smith, {U.S.\ Billion-dollar Weather and Climate Disasters, 1980--present
  (NCEI Accession 0209268)}, \url{https://doi.org/10.25921/stkw-7w73}, last
  accessed: November 2024 (2024).

\bibitem{bronstert1995}
A.~Bronstert, S.~Vollmer, J.~Ihringer, A review of the impact of land
  consolidation on runoff production and flooding in {Germany}, Physics and
  Chemistry of the Earth 20~(3--4) (1995) 321--329.

\bibitem{spitalar2014}
M.~Špitalar, et~al., Analysis of flash flood parameters and human impacts in
  the {US} from 2006 to 2012, Journal of Hydrology 519 (2014) 863--870.

\bibitem{dullo2021}
T.~T. Dullo, S.~Gangrade, M.~Morales-Hern{\'a}ndez, M.~B. Sharif, S.-C. Kao,
  A.~J. Kalyanapu, S.~Ghafoor, K.~J. Evans, Simulation of hurricane harvey
  flood event through coupled hydrologic-hydraulic models: Challenges and next
  steps, Journal of Flood Risk Management 14~(3) (2021) e12716.

\bibitem{normand2020}
A.~E. Normand, {U.S.\ Geological Survey (USGS) Streamgaging Network: Overview
  and Issues for Congress}, Tech. Rep. R45695, Congressional Research Service,
  last accessed: November 2024 (2020).

\bibitem{portalescjulia2023}
E.~Portalés-Juliá, et~al., Global flood extent segmentation in optical
  satellite images, Scientific Reports 13~(1) (2023) 20316.

\bibitem{notti2018}
D.~Notti, et~al., Potential and limitations of open satellite data for flood
  mapping, Remote Sensing 10~(11) (2018) 1673.

\bibitem{berne2004}
A.~Berne, G.~Delrieu, J.-D. Creutin, C.~Obled, Temporal and spatial resolution
  of rainfall measurements required for urban hydrology, Journal of Hydrology
  299~(3-4) (2004) 166--179.

\bibitem{9239693}
B.~Basnyat, N.~Singh, N.~Roy, A.~Gangopadhyay, Design and deployment of a flash
  flood monitoring iot: Challenges and opportunities, in: 2020 IEEE
  International Conference on Smart Computing (SMARTCOMP), 2020, pp. 422--427.
\newblock \href {https://doi.org/10.1109/SMARTCOMP50058.2020.00088}
  {\path{doi:10.1109/SMARTCOMP50058.2020.00088}}.

\bibitem{latham2025aquacam}
J.~Latham, G.~Dadzie, A.~Kalyanapu, S.~Shannigrahi, {AquaCam}: An {ML}-enhanced
  low-cost, deploy-anywhere water level detection sensor, in: 2025 IEEE
  International Conference on Pervasive Computing and Communications (PerCom),
  IEEE, 2025.

\bibitem{kalyanapu2023lowcost}
A.~Kalyanapu, C.~Owusu, T.~Wright, T.~Datta, Low-cost real-time water level
  monitoring network for {Falling Water River} watershed: A case study,
  Geosciences 13~(3) (2023) 65.

\bibitem{munoz2021tomebamba}
P.~Muñoz, J.~Orellana-Alvear, J.~Bendix, J.~Feyen, R.~Célleri, Flood early
  warning systems using machine learning techniques: The case of the
  {Tomebamba} catchment at the southern andes of {Ecuador}, Hydrology 8~(4)
  (2021) 183.
\newblock \href {https://doi.org/10.3390/hydrology8040183}
  {\path{doi:10.3390/hydrology8040183}}.

\bibitem{duncan2013rapids}
A.~P. Duncan, A.~S. Chen, E.~C. Keedwell, S.~Djordjević, D.~A. Savić,
  {RAPIDS}: Early warning system for urban flooding and water quality hazards,
  in: Proceedings of the {AISB} 2013 Symposium on Machine Learning in Water
  Systems, 2013, available at
  \url{https://ore.exeter.ac.uk/repository/handle/10871/16090}.

\bibitem{hayder2023intelligent}
I.~M. Hayder, T.~A. Al-Amiedy, W.~Ghaban, F.~Saeed, M.~Nasser, G.~A. Al-Ali,
  H.~A. Younis, An intelligent early flood forecasting and prediction
  leveraging machine and deep learning algorithms with advanced alert system,
  Processes 11~(2) (2023) 481.
\newblock \href {https://doi.org/10.3390/pr11020481}
  {\path{doi:10.3390/pr11020481}}.

\bibitem{guo2021iotlora}
T.~Guo, et~al., Design of an {IoT}-based flood early detection system using
  machine learning, in: 2021 IEEE International Conference on Smart Internet of
  Things (SmartIoT), 2021, pp. 1--6, iEEE Xplore accession 9420594.
\newblock \href {https://doi.org/10.1109/SmartIoT52359.2021.00011}
  {\path{doi:10.1109/SmartIoT52359.2021.00011}}.

\bibitem{cruz2018tencon}
F.~R.~G. Cruz, M.~G. Binag, M.~R.~G. Ga, F.~A.~A. Uy, Flood prediction using
  multi-layer artificial neural network in monitoring system with rain gauge,
  water level, soil moisture sensors, in: TENCON 2018-2018 IEEE Region 10
  Conference, IEEE, 2018, pp. 2499--2503.

\bibitem{sarangpure2022multilayer}
P.~Sarangpure, et~al., Multi-layered artificial neural network flood prediction
  system with rain gauge, temperature humidity pressure sensor, ultrasonic
  sensor, soil moisture sensor and anemometer, in: 2020 IEEE 12th International
  Conference on Humanoid, Nanotechnology, Information Technology, Communication
  and Control, Environment, and Management (HNICEM), 2020, pp. 1--6, iEEE
  Xplore accession 9072733.
\newblock \href {https://doi.org/10.1109/HNICEM51456.2020.9400099}
  {\path{doi:10.1109/HNICEM51456.2020.9400099}}.

\bibitem{mosavi2018review}
A.~Mosavi, P.~Ozturk, K.-w. Chau, Flood prediction using machine learning
  models: Literature review, Water 10~(11) (2018) 1536.

\bibitem{darkwah2024wcsna}
G.~K. Darkwah, A.~Kalyanapu, C.~Owusu, Machine learning-based flood forecasting
  system for window cliffs state natural area, {Tennessee}, GeoHazards 5~(1)
  (2024) 64--90.

\bibitem{chang2019hydroinfo}
L.-C. Chang, F.-J. Chang, S.-N. Yang, I.-F. Kao, Y.-Y. Ku, C.-L. Kuo, I.~R.
  Amin, M.~Z. bin Mat, Building an intelligent hydroinformatics integration
  platform for regional flood inundation warning systems, Water 11~(1) (2019)
  9.
\newblock \href {https://doi.org/10.3390/w11010009}
  {\path{doi:10.3390/w11010009}}.

\bibitem{tan2023algae}
Z.~Tan, et~al., A three-step machine learning approach for algal bloom
  detection using stationary {RGB} camera images, International Journal of
  Applied Earth Observation and Geoinformation 122 (2023) 103421.

\bibitem{kruger2016bridge}
A.~Kruger, W.~F. Krajewski, J.~J. Niemeier, D.~L. Ceynar, R.~Goska,
  Bridge-mounted river stage sensors ({BMRSS}), IEEE Access 4 (2016)
  8948--8966.

\bibitem{munasinghe2023multisensor}
D.~Munasinghe, et~al., A multi-sensor approach for increased measurements of
  floods and their societal impacts from space, Communications Earth \&
  Environment 4~(1) (2023) 462.

\bibitem{helmrich2021}
A.~M. Helmrich, et~al., Opportunities for crowdsourcing in urban flood
  monitoring, Environmental Modelling \& Software 143 (2021) 105124.

\bibitem{bales2009uncertainty}
J.~D. Bales, C.~R. Wagner, Sources of uncertainty in flood inundation maps,
  Journal of Flood Risk Management 2~(2) (2009) 139--147.

\bibitem{hochreiter1997lstm}
S.~Hochreiter, J.~Schmidhuber, Long short-term memory, Neural Computation 9~(8)
  (1997) 1735--1780.
\newblock \href {https://doi.org/10.1162/neco.1997.9.8.1735}
  {\path{doi:10.1162/neco.1997.9.8.1735}}.

\bibitem{kratzert2018lstm}
F.~Kratzert, D.~Klotz, C.~Brenner, K.~Schulz, M.~Herrnegger, Rainfall--runoff
  modelling using long short-term memory ({LSTM}) networks, Hydrology and Earth
  System Sciences 22~(11) (2018) 6005--6022.
\newblock \href {https://doi.org/10.5194/hess-22-6005-2018}
  {\path{doi:10.5194/hess-22-6005-2018}}.

\bibitem{fordjour2024gru}
G.~K. Fordjour, A.~J. Kalyanapu, Implementing a {GRU} neural network for flood
  prediction in {Ashland City, Tennessee}, arXiv preprint arXiv:2405.10375
  (2024).

\bibitem{teng2017review}
J.~Teng, et~al., Flood inundation modelling: A review of methods, recent
  advances and uncertainty analysis, Environmental Modelling \& Software 90
  (2017) 201--216.

\end{thebibliography}

\end{document}